\documentclass[10pt,journal,letterpaper,compsoc]{IEEEtran}
\usepackage[cmex10]{amsmath}
\usepackage{amsfonts}
\usepackage{amssymb}
\usepackage{bm}
\usepackage{graphicx}
\usepackage{graphics}
\usepackage[applemac]{inputenc}
\usepackage{cite}
\usepackage{mdwtab}
\usepackage{subfigure}
\usepackage{color}
\usepackage{xspace}
\usepackage{url}
\usepackage{tabularx}
\usepackage{epstopdf}

\usepackage{caption}

\makeatletter
\DeclareRobustCommand\onedot{\futurelet\@let@token\@onedot}
\def\@onedot{\ifx\@let@token.\else.\null\fi\xspace}

\def\eg{\emph{e.g}\onedot} 
\def\ie{\emph{i.e}\onedot} 
\def\cf{\emph{cf}\onedot} 
 \def\vs{\emph{vs}\onedot}
 
\def\etal{\emph{et al}\onedot}
\makeatother

\newcommand{\Fig}{Fig.\xspace}

\newcommand{\Sec}{Sec.\xspace}
\newcommand{\Tab}{Tab.\xspace}

\newcolumntype{L}[1]{>{\raggedright\let\newline\\\arraybackslash\hspace{0pt}}m{#1}}
\newcolumntype{C}[1]{>{\centering\let\newline\\\arraybackslash\hspace{0pt}}m{#1}}
\newcolumntype{R}[1]{>{\raggedleft\let\newline\\\arraybackslash\hspace{0pt}}m{#1}}

\newcommand{\MOTChallenge}{{\it MOTChallenge}\xspace}

\newcommand{\dismeas}{d}			
\newcommand{\simthresh}{t_d}  		

\newcommand{\parvec}{\Theta}		
\newcommand{\onepar}{\theta}		

\newcommand{\Tracker}[1]{\textsc{\lowercase{#1}}\xspace}

\graphicspath{{figures/},{figures/moving/},{figures/static/},{figures/annotations/}}

\definecolor{darkgreen}{rgb}{0,.75,0}
\definecolor{gray40}{gray}{.40}

   \teaser{
   \centering
   \hspace{0.005\linewidth}
   \includegraphics[width=.92\linewidth]{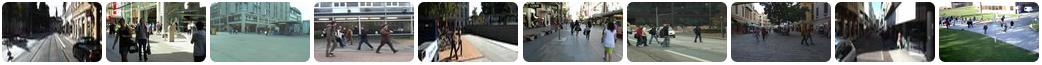}
  }

\begin{document}
\title{MOTChallenge 2015: \\Towards a Benchmark for Multi-Target Tracking}

\author{Laura Leal-Taix{\'e}$^*$,
        Anton Milan$^*$,
        Ian Reid, 
        Stefan Roth,
        and Konrad Schindler
\thanks{$^*$ = authors contributed equally.}
\thanks{L.~Leal-Taix{\'e}$^*$ and K.~Schindler are with the Photogrammetry and Remote Sensing Group at ETH Zurich, Switzerland.}
\thanks{A.~Milan$^*$ and I.~Reid are with the Australian Centre for Visual Technologies at the University of Adelaide, Australia.}
\thanks{S.~Roth is with the Department of Computer Science, Technische Universit{\"a}t Darmstadt, Germany.}
\thanks{Primary contacts: leal@geod.baug.ethz.ch, anton.milan@adelaide.edu.au }
}

\IEEEcompsoctitleabstractindextext{%

\begin{abstract}

In the recent past, the computer vision community has developed
centralized benchmarks for the performance evaluation of a variety of
tasks, including generic object and pedestrian detection, 3D
reconstruction, optical flow, single-object short-term tracking, and
stereo estimation.
Despite potential pitfalls of such benchmarks, they have proved to be
extremely helpful to advance the state of the art in the respective
area. Interestingly, there has been rather limited work on
the standardization of quantitative benchmarks for multiple target
tracking.
One of the few exceptions is the well-known PETS dataset
\cite{Ferryman:2010:PETS}, targeted primarily at surveillance
applications.
Despite being widely used, it is often applied inconsistently,
for example involving using different subsets of the available data,
different ways of training the models, or differing evaluation scripts.
This paper describes our work toward a novel multiple object tracking
benchmark aimed to address such issues.
We discuss the challenges of creating such a framework, collecting
existing and new data, gathering state-of-the-art methods to be tested
on the datasets, and finally creating a unified evaluation system.
With \MOTChallenge we aim to pave the way toward a unified
evaluation framework for a more meaningful quantification of
multi-target tracking.

\end{abstract}

}

\maketitle

\IEEEdisplaynotcompsoctitleabstractindextext

\IEEEpeerreviewmaketitle


\section{Introduction}
\label{sec:introduction}

Evaluating and comparing multi-target tracking methods is not trivial 
for numerous reasons (\emph{cf. e.g.} \cite{Milan:2013:CVPRWS}). 
First, unlike for other tasks, such as image denoising, the ground 
truth, \emph{i.e.} the perfect solution one aims to achieve, is
difficult to define clearly. Partially visible, occluded, or cropped targets, 
reflections in mirrors or windows, and objects that very closely resemble 
targets all impose intrinsic ambiguities, such that even humans may 
not agree on one particular ideal solution. Second, a number of different
evaluation metrics with free parameters and ambiguous definitions often 
lead to inconsistent quantitative results across the literature. Finally, the 
lack of pre-defined test and training data makes it difficult to compare
different methods fairly.

Multi-target tracking is a crucial problem in scene understanding, which, 
in contrast to other research areas, still lacks large-scale benchmarks. 
We believe that a unified evaluation platform that allows participants to submit 
not only their own tracking methods, but also their own data, including 
video and annotations, as well as propose new evaluation methodologies that can 
be applied instantaneously to all previous approaches, can bring a vast 
benefit to the computer vision community. 

To that end we develop the \MOTChallenge benchmark, consisting of
three main components:
(1) a publicly available dataset, (2) a centralized evaluation method, and (3) 
an infrastructure that allows for crowdsourcing of new data, new 
evaluation methods and even new annotations. The first release of the dataset contains a 
total of 22 sequences, half for training and half for testing, with a total 
of 11286 frames or 996 seconds of video. Camera calibration is provided 
for 4 of those sequences to enable 3D real-world coordinate tracking. We 
also provide pre-computed object detections, annotations, and a common evaluation method for 
all datasets, so that all results can be compared in a fair way. The 
final goal is to collect sequences from all researchers who are willing 
to contribute to the benchmark, enabling an update of the data, new 
evaluation tools, new annotations, etc.~to be made available yearly. 
%

We anticipate two ways of submitting to the \MOTChallenge benchmark: (1) 
year-round, or (2) submissions to a specific workshop or challenge, which 
is to be held once a year. The purpose of the former is to keep track
of state-of-the-art methods submitted at major conferences and
journals, to allow for a fair comparison between methods by ensuring
that all are using the same datasets and the same evaluation methods.
The latter follows the well-known format of yearly challenges that have
been very successful in other areas, \emph{e.g.}, in the PASCAL VOC 
series \cite{Everingham:2012:VOC} or the ImageNet competitions 
\cite{Russakovsky:2014:arxiv}. These challenges and workshops provide a 
way to track and discuss the progress and innovations of 
state-of-the-art methods presented over the year. The first workshop 
\cite{bmtt2015} organized on the \MOTChallenge benchmark took place in 
early 2015 in conjunction with the Winter Conference on Applications of 
Computer Vision (WACV).
\smallskip

\noindent{\bf Goals.} This paper has three main goals:

\begin{enumerate}
 \item To discuss the challenges of creating a multi-object tracking 
benchmark;
 \item to analyze current datasets and evaluation methods;

  \item to bring forward the strengths and weaknesses of state-of-the-art 
multi-target tracking methods.
 
\end{enumerate}

The benchmark with all datasets, current ranking and submission 
guidelines can be found at:
\begin{center}
\url{http://www.motchallenge.net/}
\end{center}

\subsection{Related work}
\label{sec:related-work}
\noindent{\bf Benchmarks and challenges.}
In the recent past, the computer vision community has developed
centralized benchmarks for numerous tasks including object detection 
\cite{Everingham:2012:VOC}, pedestrian detection 
\cite{caltechpedestrians}, 3D reconstruction \cite{Seitz:2006:CVPR}, 
optical flow \cite{Baker:2011:IJCV,Geiger:2012:CVPR}, visual odometry 
\cite{Geiger:2012:CVPR}, single-object short-term tracking 
\cite{VOC2014}, and stereo estimation \cite{Scharstein:2002:IJCV, 
Geiger:2012:CVPR}. Despite potential pitfalls of such benchmarks (\eg~\cite{Torralba:2011:ULD}), they 
have proved to be extremely helpful to advance the state of the art in 
the respective area. For multiple target tracking, in contrast, there
has been very limited work on standardizing quantitative
evaluation.

One of the few exceptions is the well known PETS dataset 
\cite{Ferryman:2010:PETS}, targeted primarily at surveillance 
applications. The 2009 version consisted of 3 subsets, S1 targeted 
at person count and density estimation, S2 targeted at people 
tracking, and S3 targeted at flow analysis and event recognition. The 
easiest sequence for tracking (S2L1) consisted of a scene with few pedestrians, 
and for that sequence state-of-the-art methods perform extremely well 
with accuracies of over 90\% given a good set of initial detections 
\cite{Milan:2014:PAMI, Henriques:2011:ICCV, Zamir:2012:ECCV}. Methods then 
moved to tracking on the hardest of the sequences (with most crowd
density) , but hardly ever on the complete dataset. Even for this
widely used benchmark, we observe that tracking results are commonly
obtained in an inconsistent fashion: involving using different subsets of 
the available data, different detection inputs, inconsistent model 
training that is often prone to overfitting, and varying evaluation
scripts.
Results are thus not easily comparable.  
So the question that arises here is: Are these sequences already too easy 
for current tracking methods, are methods simply overfitted, or are
they poorly evaluated?

The PETS team organized a workshop approximately once a year to which
researchers could submit their results, and methods were evaluated under 
the same conditions. Although this was indeed a fair comparison, the 
fact that submission was only once a year meant that the use of this 
benchmark for high impact conferences like ICCV or CVPR still remained 
an issue. 

A well-established and useful way of organizing datasets is through 
standardized challenges. These are usually in the form of web servers 
that host the data and through which results are uploaded by the
users.
Results are then computed in a centralized way by the server and
afterwards presented online to the public, making comparison with any
other method immediately possible.
There are several datasets organized in this fashion: the Labeled Faces 
in the Wild \cite{huangtech2007} for unconstrained face recognition, the 
PASCAL VOC \cite{Everingham:2012:VOC} for object detection, the ImageNet 
large scale visual recognition challenge \cite{Russakovsky:2014:arxiv},
or the Reconstruction Meets Recognition Challenge (RMRC) \cite{rmrc2014}.


Recently, the KITTI benchmark \cite{Geiger:2012:CVPR} was introduced for 
challenges in autonomous driving, which included stereo/flow, odometry, 
road and lane estimation, object detection and orientation estimation, as 
well as tracking. Some of the sequences include crowded pedestrian 
crossings, making the dataset quite challenging, but the camera position 
is always the same for all sequences (at a car's height). 

With the \MOTChallenge benchmark, we aim to increase the difficulty 
by including a variety of sequences filmed from different 
viewpoints, with different lighting conditions, and different levels of 
crowd density. In addition to other existing and new data, we include 
sequences from both PETS and KITTI datasets. The real challenge we see
is not to perform well on an individual sequence, but rather to perform well 
on a diverse set of sequences.



Another work that is worth mentioning is \cite{alahicvpr2014}, in which 
the authors collect a very large amount of data with 42 million pedestrian 
trajectories. Since annotation of such a large collection of data is 
infeasible, they use a denser set of cameras to create the ``ground 
truth'' trajectories. Though we do not aim at collecting such a large 
amount of data, the goal of our benchmark is somewhat similar: to push research 
in tracking forward by generalizing the test data to a larger set that is 
highly variable and hard to overfit. 
\bigskip


\noindent{\bf Evaluation.} 
A critical point with any dataset is how to measure the performance of 
the algorithms. In the case of multiple object tracking, the CLEAR 
metrics \cite{clear} have emerged as one of the standard measures. By 
measuring the intersection over union of bounding boxes and matching those 
from annotations and results, measures of accuracy and precision can be 
computed. Precision measures how well the persons are localized, while 
accuracy evaluates how many distinct errors such as missed targets, ghost
trajectories, or identity switches are made.

Another set of measures that is widely used in the tracking community is 
that of \cite{Li:2009:CVPR}. There are three widely used metrics 
introduced in that work: mostly tracked, mostly lost, and partially
tracked pedestrians. These numbers give a very good intuition on the  
performance of the method. We refer the reader to 
Section~\ref{sec:evaluation} for more formal definitions.

A key parameter in the both families of metrics is the intersection-over-union 
threshold, which determines if a bounding box is matched to an 
annotation or not. It is fairly common to observe methods compared under 
different thresholds, varying from 25\% to 50\%. There are often many other 
variables and implementation details that differ between 
evaluation scripts, but which may affect results significantly. 
%

It is therefore clear that standardized benchmarks are the only way to 
compare methods in a fair and principled way. Using the same ground 
truth data and evaluation methodology is the only way to 
guarantee that the only part being evaluated is the tracking method that 
delivers the results. This is the main goal behind this paper and behind 
the \MOTChallenge benchmark.

\section{Benchmark Submission}
\label{sec:benchmark-structure}
Our benchmark consists of the database and evaluation server on one hand, 
and the website as the user interface on the other. It is open to 
everyone who respects the submission policies (see next section). Before 
participating, every user is required to create an account, if possible
providing an institutional and not a generic e-mail
address\footnote{For accountability and to prevent abuse by using several email accounts.}. 
After registering, the user can create a 
new tracker with a unique name and enter all additional details. It is 
mandatory to indicate
\begin{itemize}
\item the challenge or benchmark category in which the tracker will be 
participating,
\item the full name and a brief description of the method including the parameters used,

\item whether the method operates online or on a 
batch of frames and whether the source code is publicly available,

\item the \emph{total} runtime in seconds for computing the results on all 
sequences and the hardware used, and 
\item whether only the provided or also external training and detection 
data were used. 

\end{itemize}
After entering all details, it is possible to submit the results in 
the format described in \Sec~\ref{sec:data-format}. The tracking 
results will be automatically evaluated and appear on the user's profile. 
They will \emph{not} be displayed in the public ranking table. 
The user can then decide at any point in time to 
make the results public. Note that the results can be published anonymously,
\eg to enable a blind review process for a corresponding paper. In this case,
we ask to provide the venue and the paper ID or a similar unique reference.
We request that a proper reference to the method's description is added 
upon acceptance of the paper. In case of rejection, an anonymous entry may 
also be removed from the database. Anonymous entries will also be removed
after six months of inactivity.

The tracker's meta information such as description, or project page can be edited
at any time. Visual results of all public submissions, as well as annotations and
detections can be viewed on a dedicated 
visualization page\footnote{\url{http://motchallenge.net/vis/}}.

\subsection{Submission policy}
The main goal of this benchmark is to provide a platform that allows for 
an objective performance comparison of multiple target tracking 
approaches on real-world data. Therefore, we introduce a few simple 
guidelines that must be followed by all participants.

{\bf Training.} Ground truth is only provided for the training sequences.
It is the participant's own responsibility to find the best setting
using \emph{only} the training data.
The use of additional training data must be indicated during submission and will
be visible in the public ranking table.
The use of ground truth labels on the test data is strictly forbidden. This or any other misuse
of the benchmark will lead to the deletion of the participant's account and their results.

{\bf Detections.} We also provide a unique set of detections 
(see \Sec \ref{sec:detections}) for each sequence. We expect all tracking-by-detection
algorithms to use the given detections. In case the user wants to present results with another set of
detections or is not using detections at all, this should be clearly stated during submission
and will also be displayed in the results table.

{\bf Submission frequency.} Generally, we expect one single submission 
for every tracking approach. If for any reason, the user needs to 
re-compute and re-submit the results (\eg due to a bug discovered in the 
implementation), he/she may do so after a waiting period of 72 hours 
after the last submission. This policy should discourage the use of the
benchmark server for training and parameter tuning on the test data. The number of
submissions is counted and displayed for each method.
Under \emph{no} circumstances must anyone create a second account and 
attempt to re-submit in order to bypass the waiting period. This may lead
to a deletion of the account and excluding the user from participating
in the benchmark.

\subsection{Challenges}
Besides the main benchmarks (2D MOT 2015, 3D MOT 2015), we anticipate to organize 
multi-target tracking challenges on a regular basis, similar in spirit 
to the widely known PASCAL VOC series \cite{Everingham:2012:VOC}, or the 
ImageNet competitions \cite{Russakovsky:2014:arxiv}. The main differences
to the main benchmark are:
\begin{itemize}
\item The dataset is typically smaller, but potentially more challenging.
\item There is a fixed submission deadline for all participants.
\item The results are revealed and the winners awarded at a 
corresponding workshop.
\end{itemize}
The first edition of our series was the WACV 2015 Challenge that 
consisted of six new outdoor sequences with both moving and static cameras. 
The results were presented 
at the BMTT Workshop \cite{bmtt2015} held in conjunction with WACV 2015.

\section{Datasets}
\label{sec:datasets}

\begin{table*}[tbp]
\begin {center}
 \begin{tabular}{|l| c| c| c| c| c| c| c |c | c | c| c|}
 \hline
 \multicolumn{12}{|c|}{\bf Training sequences} \\ 
 \hline 
      Name & FPS & Resolution & Length & Tracks & Boxes & Density & 3D & Camera & Viewpoint & Shadows & Source\\ 
      \hline
     TUD-Stadtmitte & 25 & 640x480 & 179 (00:07) & 10 & 1156 & 6.5 & yes & static  & medium & cloudy & \cite{Andriluka:2010:CVPR} \\
     TUD-Campus & 25 & 640x480 & 71 (00:03) & 8 & 359 & 5.1 & no & static & medium & cloudy & \cite{Andriluka:2008:CVPR} \\
    PETS09-S2L1 & 7 & 768x576 & 795 (01:54) & 19 & 4476 & 5.6 & yes & static & high & cloudy & \cite{Ferryman:2010:PETS} \\
    ETH-Bahnhof & 14 & 640x480 & 1000 (01:11) & 171 & 5415 & 5.4 & yes & moving & low & cloudy	& \cite{Ess:2008:CVPR} \\
    ETH-Sunnyday & 14 & 640x480 & 354 (00:25) & 30 & 1858 & 5.2 & yes & moving & low & sunny	& \cite{Ess:2008:CVPR} \\
     ETH-Pedcross2 & 14 & 640x480 & 840 (01:00) & 133 & 6263 & 7.5 &no & moving & low & sunny & \cite{Ess:2008:CVPR} \\
     ADL-Rundle-6 & 30 & 1920x1080 & 525 (00:18) & 24 & 5009 & 9.5 &no & static & low & cloudy & new \\
      ADL-Rundle-8 & 30 & 1920x1080 & 654 (00:22) & 28 & 6783 & 10.4 &no & moving & medium & night & new \\
      KITTI-13 & 10 & 1242x375 & 340 (00:34) & 42 & 762 & 2.2 &no & moving & medium & sunny & \cite{Geiger:2012:CVPR} \\
       KITTI-17 & 10 & 1242x370 & 145 (00:15) & 9 & 683 & 4.7 &no & static & medium & sunny & \cite{Geiger:2012:CVPR} \\
      Venice-2 & 30 & 1920x1080 & 600 (00:20) & 26 & 7141 & 11.9 &no & static & medium & sunny & new\\
      \hline
      \multicolumn{3}{|c|}{\bf Total training} & {\bf 5503 (06:29)} & {\bf 500} & {\bf 39905} & {\bf 7.3} & & & & &  \\
    \hline
    \multicolumn{12}{c}{\vspace{1em}} \\
    \hline
     \multicolumn{12}{|c|}{\bf Testing sequences} \\ 
     \hline 
 Name & FPS & Resolution & Length & Tracks & Boxes & Density & 3D &  Camera & Viewpoint & Weather & Source\\ 
 \hline
     TUD-Crossing & 25 & 640x480 & 201 (00:08) & 13 & 1102 & 5.5 & no & static  & medium & cloudy & \cite{Andriluka:2008:CVPR} \\   
     PETS09-S2L2 & 7 & 768x576 & 436 (01:02) & 42 & 9641 & 22.1 & yes & static & high & cloudy & \cite{Ferryman:2010:PETS} \\
     ETH-Jelmoli & 14 & 640x480 & 440 (00:31) & 45 & 2537 & 5.8 & yes & moving & low & sunny	& \cite{Ess:2008:CVPR} \\
    ETH-Linthescher & 14 & 640x480 & 1194 (01:25) & 197 & 8930 & 7.5 & yes & moving & low & sunny	& \cite{Ess:2008:CVPR} \\
     ETH-Crossing & 14 & 640x480 & 219 (00:16) & 26 & 1003 & 4.6 &no & moving & low & cloudy & \cite{Ess:2008:CVPR} \\
     AVG-TownCentre & 2.5 & 1920x1080 & 450 (03:45) & 226 & 7148 & 15.9 & yes & static & high & cloudy & \cite{Benfold:2011:CVPR}\\
     ADL-Rundle-1 & 30 & 1920x1080 & 500 (00:17) & 32 & 9306 & 18.6 &no & moving & medium & sunny & new \\
      ADL-Rundle-3 & 30 & 1920x1080 & 625 (00:21) & 44 & 10166 & 16.3 &no & static & medium & sunny & new \\
      KITTI-16 & 10 & 1242x370 & 209 (00:21) & 17 & 1701 & 8.1 &no & static & medium & sunny& \cite{Geiger:2012:CVPR} \\
       KITTI-19 & 10 & 1242x374 & 1059 (01:46) & 62 & 5343 & 5.0 &no & moving & medium & sunny & \cite{Geiger:2012:CVPR} \\ 
       Venice-1 & 30 & 1920x1080 & 450 (00:15) & 17 & 4563 & 10.1 & no & static & medium & sunny & new \\ 
       \hline
      \multicolumn{3}{|c|}{\bf Total testing} & {\bf 5783 (10:07)} & {\bf 721} & {\bf 61440} & {\bf 10.6} & & & & & \\
      \hline 
    \end{tabular}
  \end{center}
    \caption{Overview of the sequences currently included in the benchmark.}
\label{tab:dataoverview}
\end{table*}
One of the key aspects of any benchmark is data collection. The goal of \MOTChallenge is 
not only to compile yet another dataset with completely new data, but rather 
to: (1) create a common framework to test tracking methods on;  
%
(2) gather existing and new challenging sequences 
with very different characteristics (frame rate, 
pedestrian density, illumination or point of view) in order to challenge 
researchers to develop more general tracking methods that can deal with all 
types of sequences.
In Table \ref{tab:dataoverview} we show an overview of the sequences 
included in the benchmark.

\subsection{2D MOT 2015 sequences}

We have compiled a total of 22 sequences, of which we use half for 
training and half for testing. The annotations of the testing sequences 
will not be released in order to avoid (over)fitting of the methods to the 
specific sequences. Nonetheless, the test data contains over 10 minutes 
of footage and 61440 annotated bounding boxes, therefore, it is hard for 
algorithms to overtune on such a large amount of data. This is one of the 
major strengths of the benchmark. 

Sequences are very different from each other, we can classify them according to: \\

\begin{itemize}
 \item Moving or static camera: the camera can be held by a person \cite{bmtt2015}, placed on a stroller \cite{Ess:2008:CVPR} or on a car \cite{Geiger:2012:CVPR}, or can be positioned fixed in the scene.
 \item Viewpoint: the camera can overlook the scene from a high position, a medium position (at pedestrian's height), or at a low position. 
 \item Weather: the weather conditions in which the sequence was taken are reported in order to get an estimate of the illumination conditions of the scene. Sunny sequences may contain shadows and saturated parts of the image, while the night sequence contains a lot of motion blur, making pedestrian detection and tracking rather challenging. Cloudy sequences on the other hand contain fewer of those artifacts.\\
\end{itemize}

\begin{figure*}[htpb]
\centering
\subfigure[Camera motion]{
\includegraphics[trim=3.5cm 0.5cm 2.5cm 1cm,clip=true,width=0.31\linewidth]{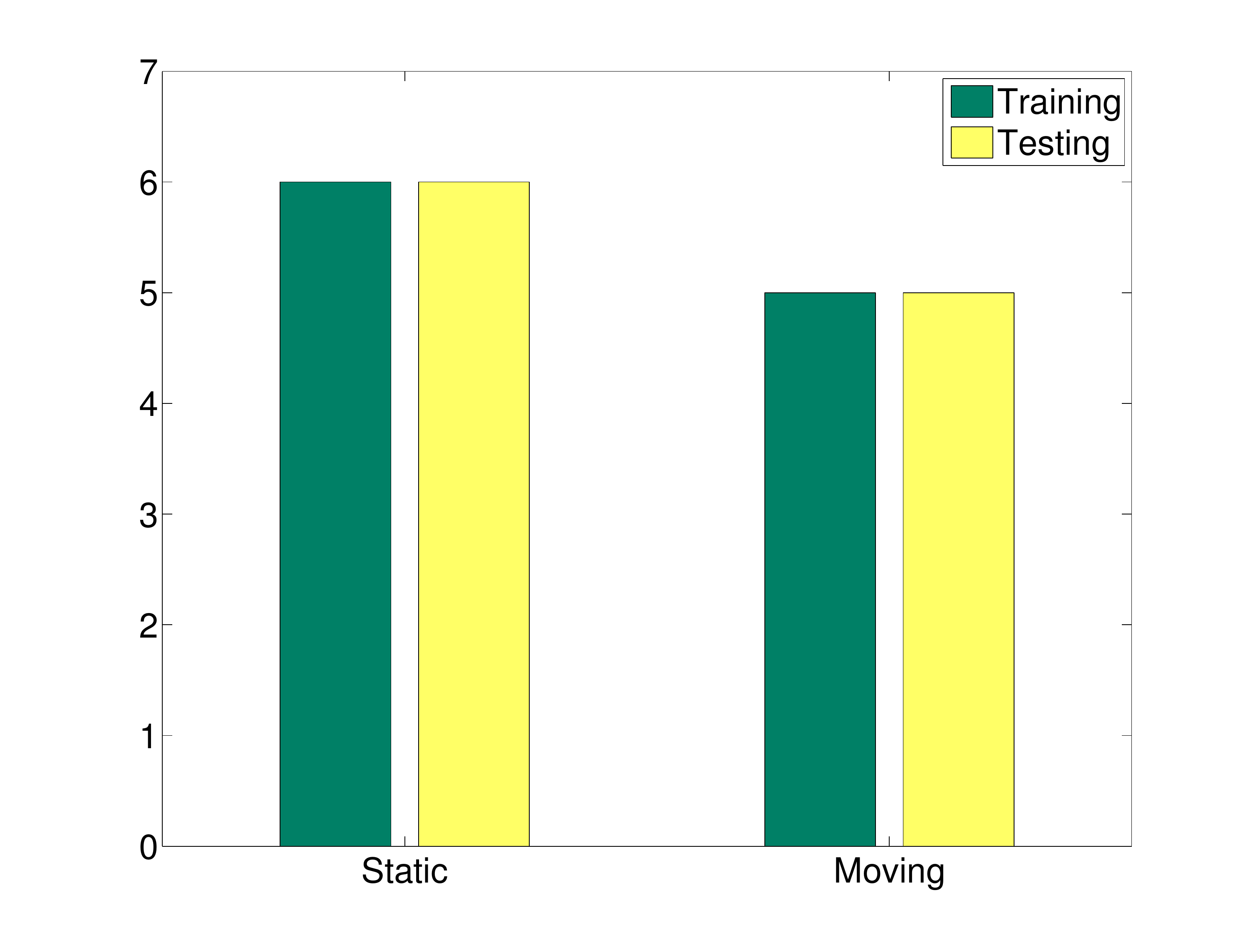} 
\label{fig:camera}}
\subfigure[Viewpoint]{
\includegraphics[trim=3.5cm 0.5cm 2.5cm 1cm,clip=true,width=0.31\linewidth]{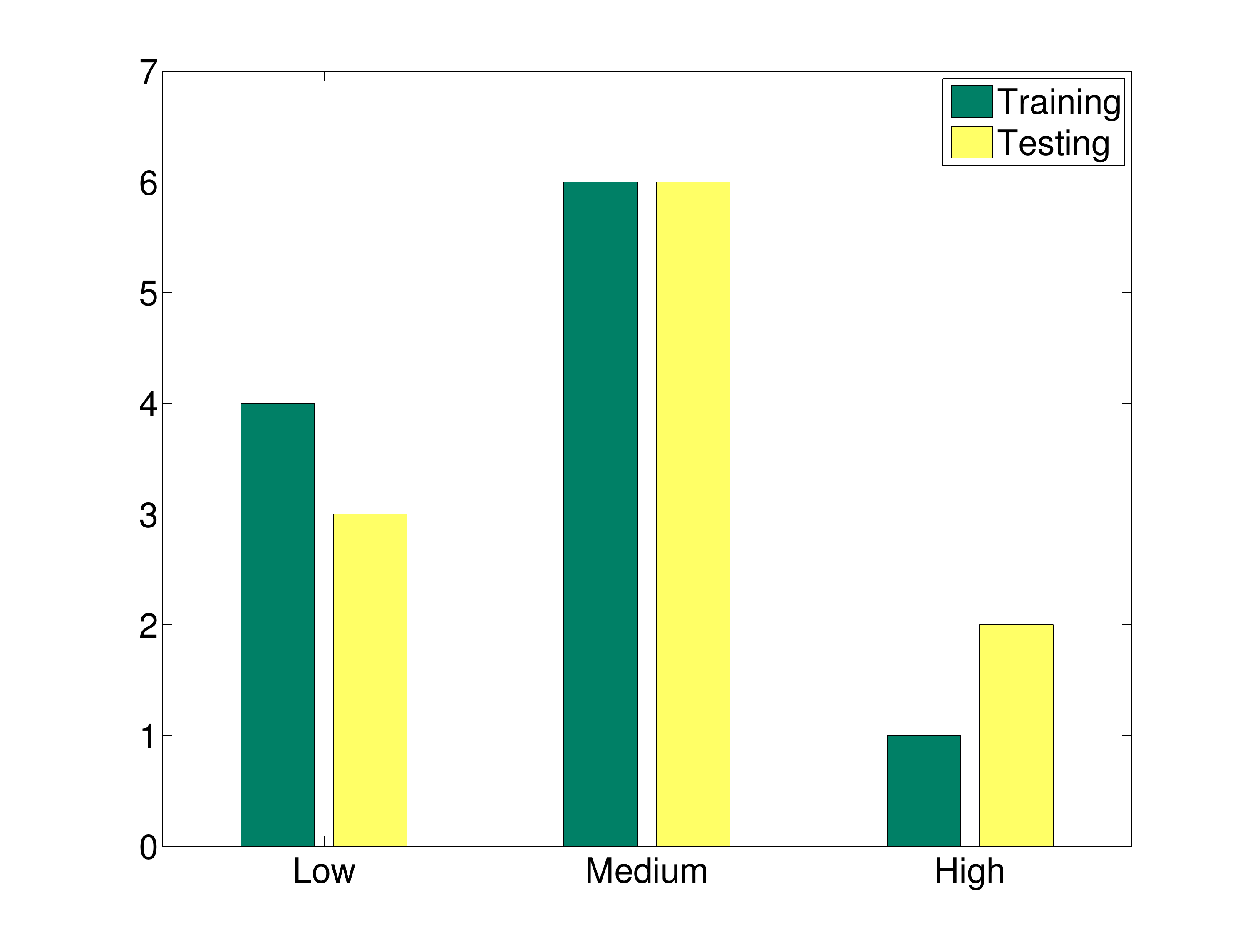} 
\label{fig:viewpoint}}
\subfigure[Weather conditions]{
\includegraphics[trim=3.5cm 0.5cm 2.5cm 1cm,clip=true,width=0.31\linewidth]{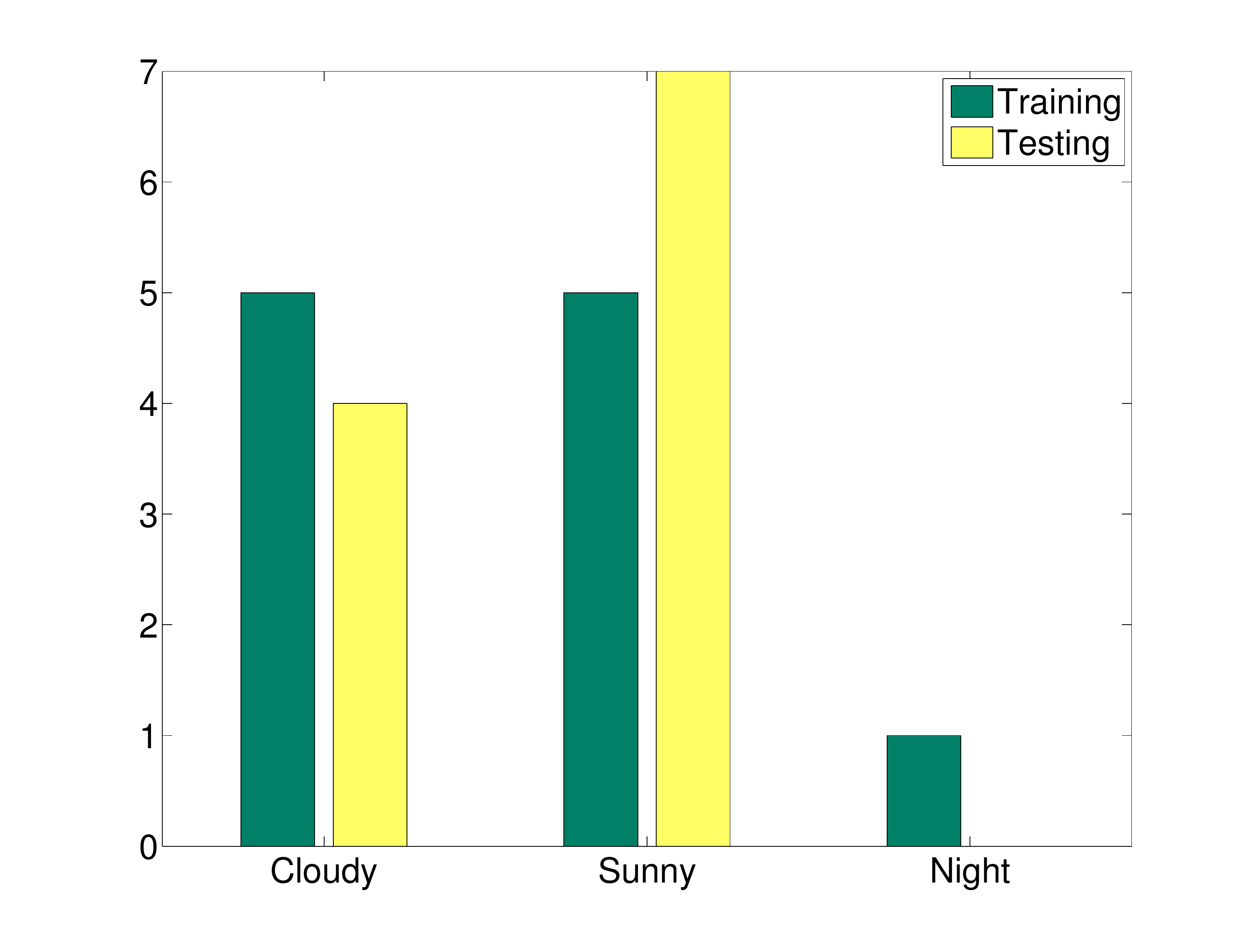} 
\label{fig:weather}}
\caption{Comparison histogram between training and testing sequences of (a) static vs. moving camera, (b) camera viewpoint: low, medium or high, (c) weather conditions: cloudy, sunny, or night.}
\label{fig:2Ddata}
\end{figure*}

We divided the sequences into training and testing in order to have a balanced distribution, as we can see in Figure \ref{fig:2Ddata}.

\subsubsection{New sequences}

We introduce 6 new challenging sequences, 4 filmed from a static camera 
and 2 from a moving camera held at pedestrian's height. Three of the 
sequences are particularly difficult: a night sequence filmed from a moving camera and two outdoor sequences with a high density of pedestrians. The moving 
camera together with the low illumination creates a lot of motion blur, 
making this sequence extremely challenging. %
In the future, we will include further sequences captured on rainy or foggy days and evaluate how 
methods perform under those special conditions.
A special challenge including only these 6 new sequences was held at the 1st 
Workshop on Benchmarking Multi-Target Tracking \cite{bmtt2015}. The best 
performing algorithm reached a MOTA (tracking accuracy) of 12.7\%, 
showing how challenging these new sequences are\footnote{The challenge results are available at \url{http://motchallenge.net/results/WACV_2015_Challenge/}.}.


\subsection{3D MOT 2015 sequences}

A pedestrian's 3D position is typically obtained by projecting the 2D position of the feet of the person 
into the 3D world, \eg, by using a homography between the image plane and the ground plane.
The bottom-center point of the bounding box is commonly chosen to
represent the position of the feet of the pedestrian, but this may
not be particularly accurate, if the bounding box is not placed very tightly around the pedestrian's silhouette, or if the limbs are extended asymmetrically. By the nature of projective geometry, even slight 2D misplacements can cause large 3D errors. 
It is therefore clear, that obtaining accurate 3D information only from bounding boxes is a challenging task. 

In this section, we detail how the 3D information is obtained for both static and moving cameras, and
discuss whether current calibration and annotation pairs are accurate enough for reliable 3D tracking.
For the sequences with a moving
camera we show that these errors are too large for tracking purposes,
and therefore we argue not to include those sequences in the 3D benchmark.
We thus limit the 3D category to a few available 3D sequences with a static camera, but plan
to extend the number of 3D sequences in the near future.

\subsubsection{Static camera sequences}
\begin{figure*}[htpb]
\centering
\subfigure[PETS09-S2L1]{
\includegraphics[trim=2.0cm 0.5cm 2.4cm 0.9cm,clip=true,width=0.23\linewidth]{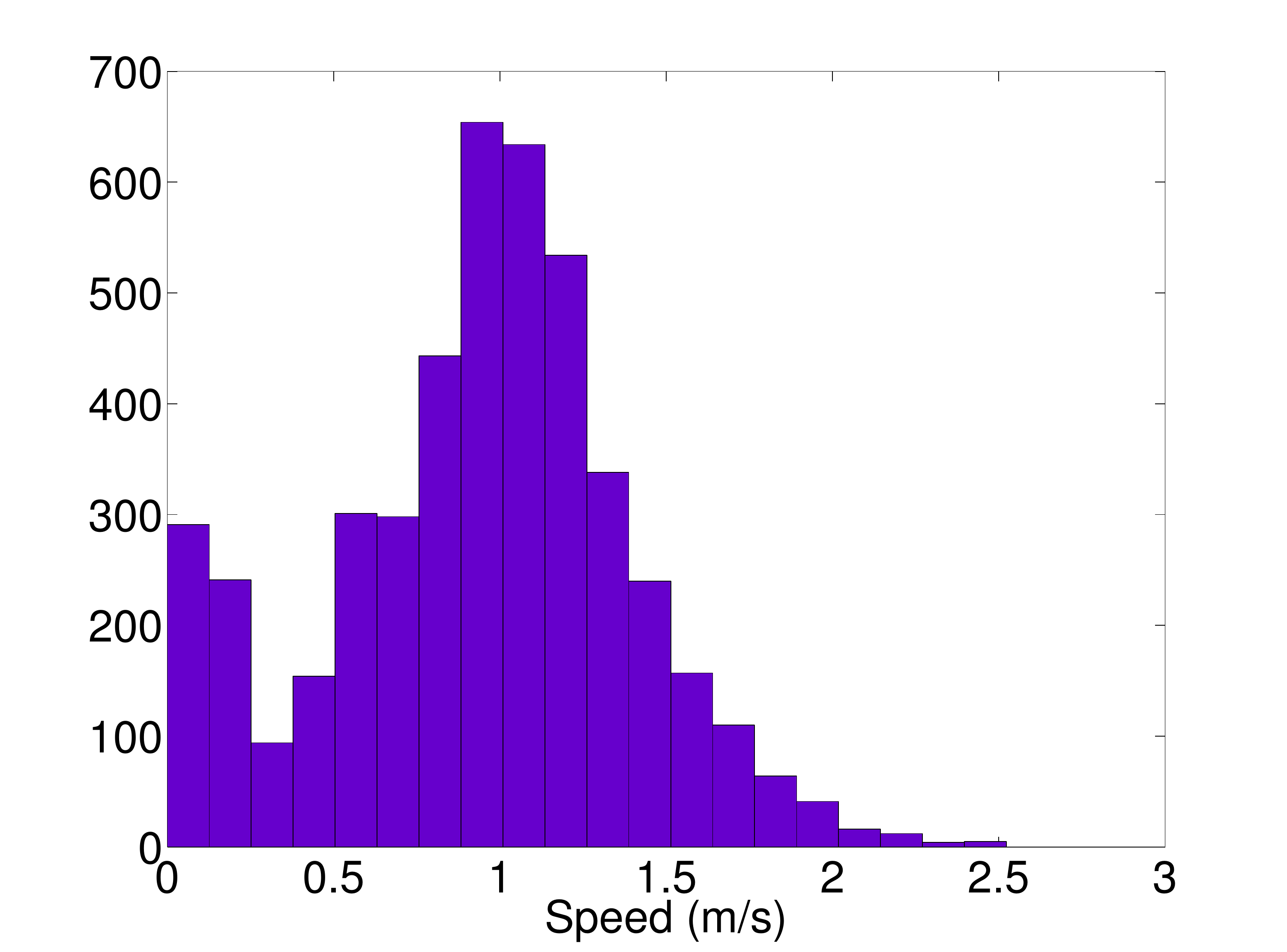} 
\label{fig1s}}
\subfigure[PETS09-S2L2]{
\includegraphics[trim=2.0cm 0.5cm 2.4cm 0.9cm,clip=true,width=0.23\linewidth]{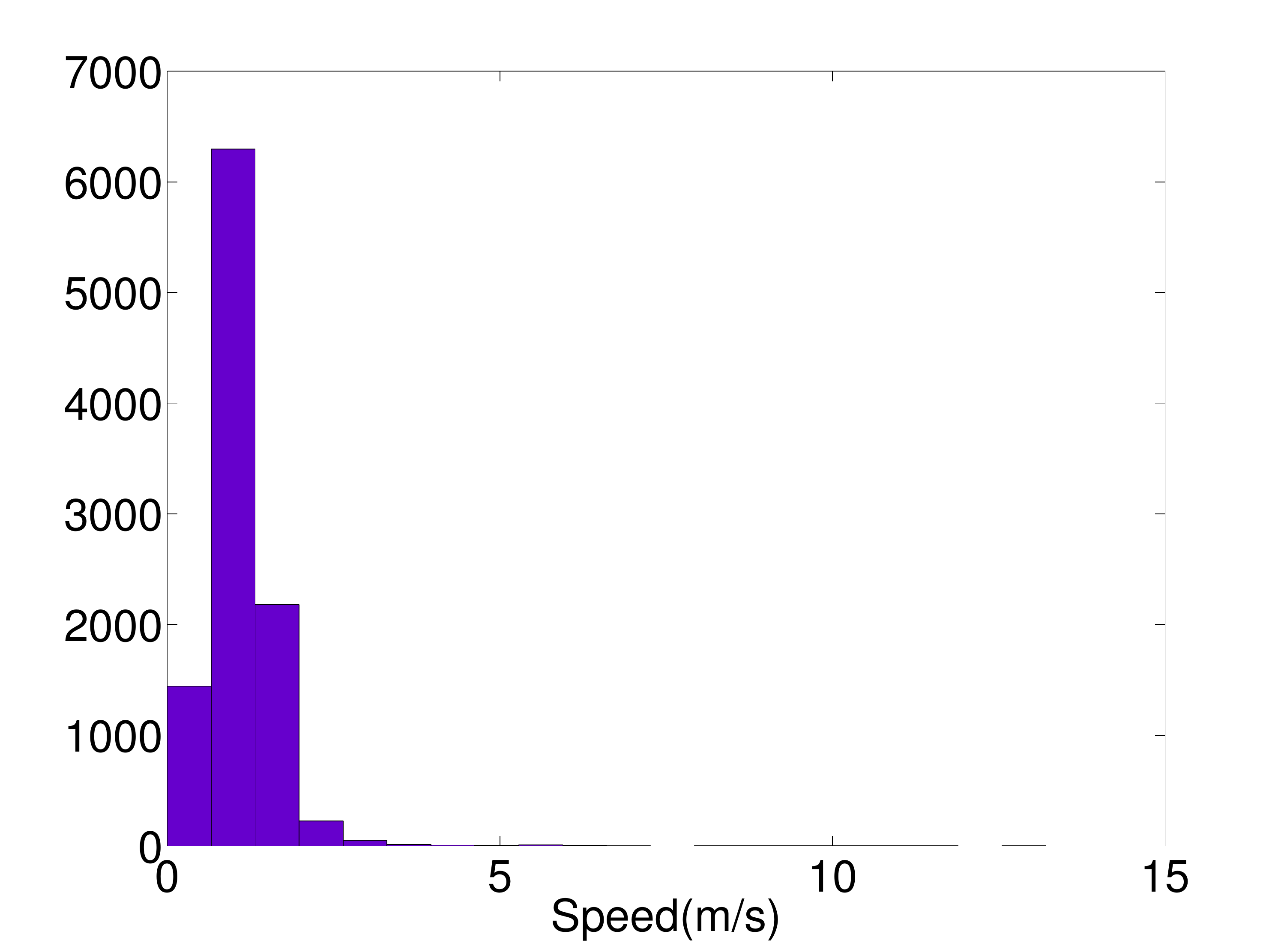} 
\label{fig2s}}
\subfigure[AVG-TownCentre]{
\includegraphics[trim=2.0cm 0.5cm 2.4cm 0.9cm,clip=true,width=0.23\linewidth]{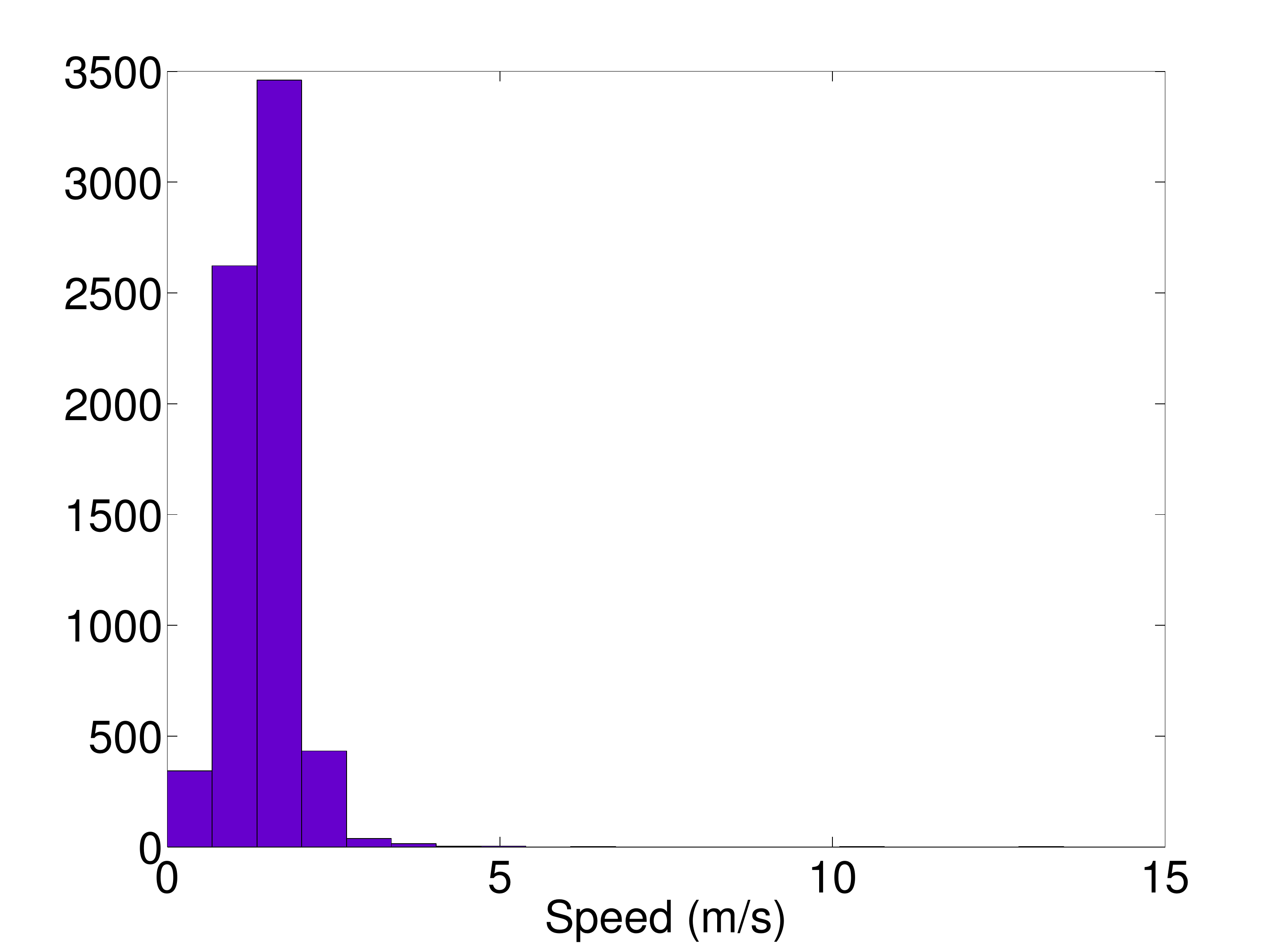} 
\label{fig3s}}
\subfigure[TUD-Stadtmitte]{
\includegraphics[trim=2.0cm 0.5cm 2.4cm 0.9cm,clip=true,width=0.23\linewidth]{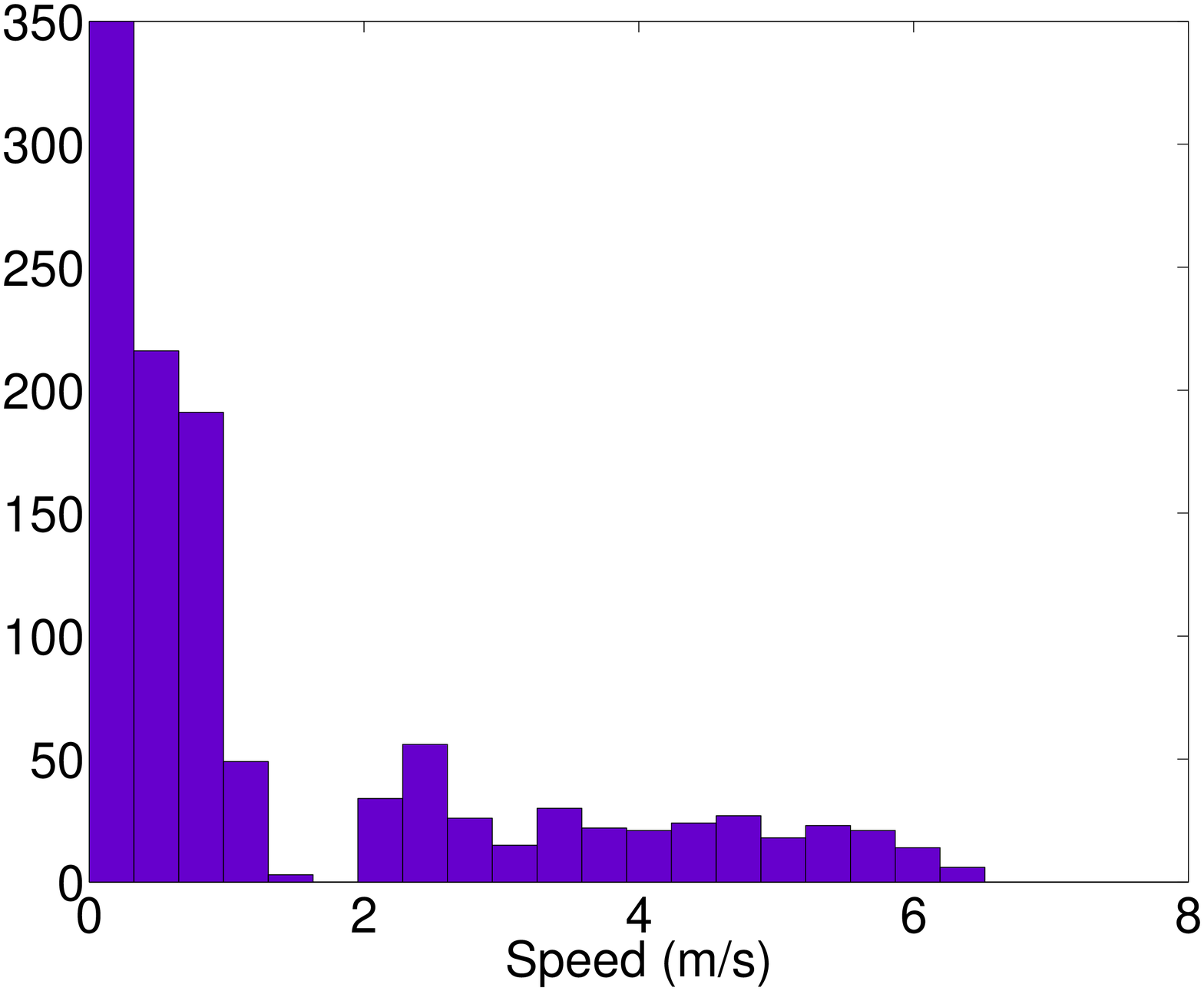} 
\label{fig4s}}
\subfigure[PETS09-S2L1]{
\includegraphics[trim=1.0cm 0.5cm 2.2cm 0.9cm,clip=true,width=0.23\linewidth]{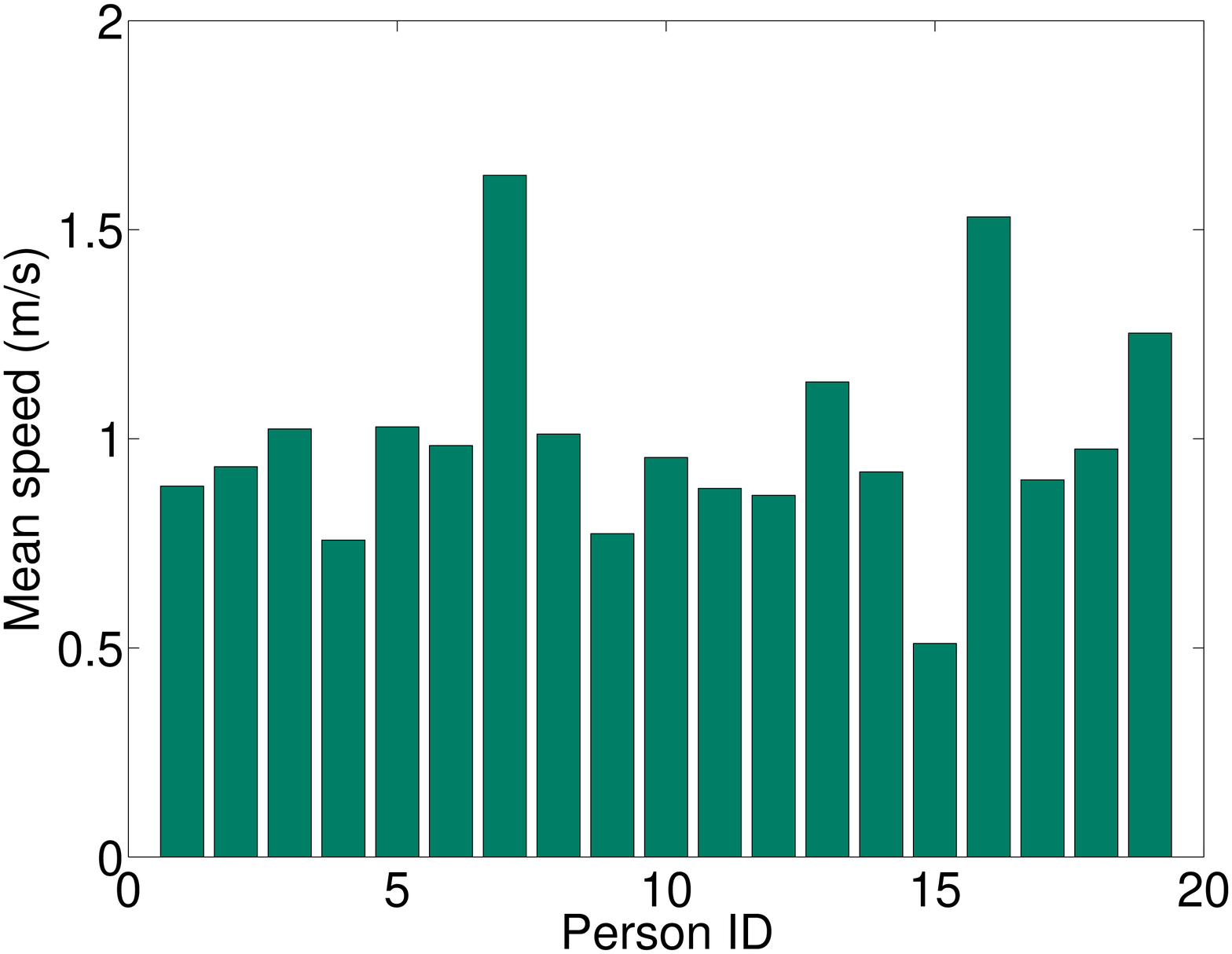} 
\label{fig5s}}
\subfigure[PETS09-S2L2]{
\includegraphics[trim=1.0cm 1.55cm 2.2cm 0.9cm,clip=true,width=0.241\linewidth]{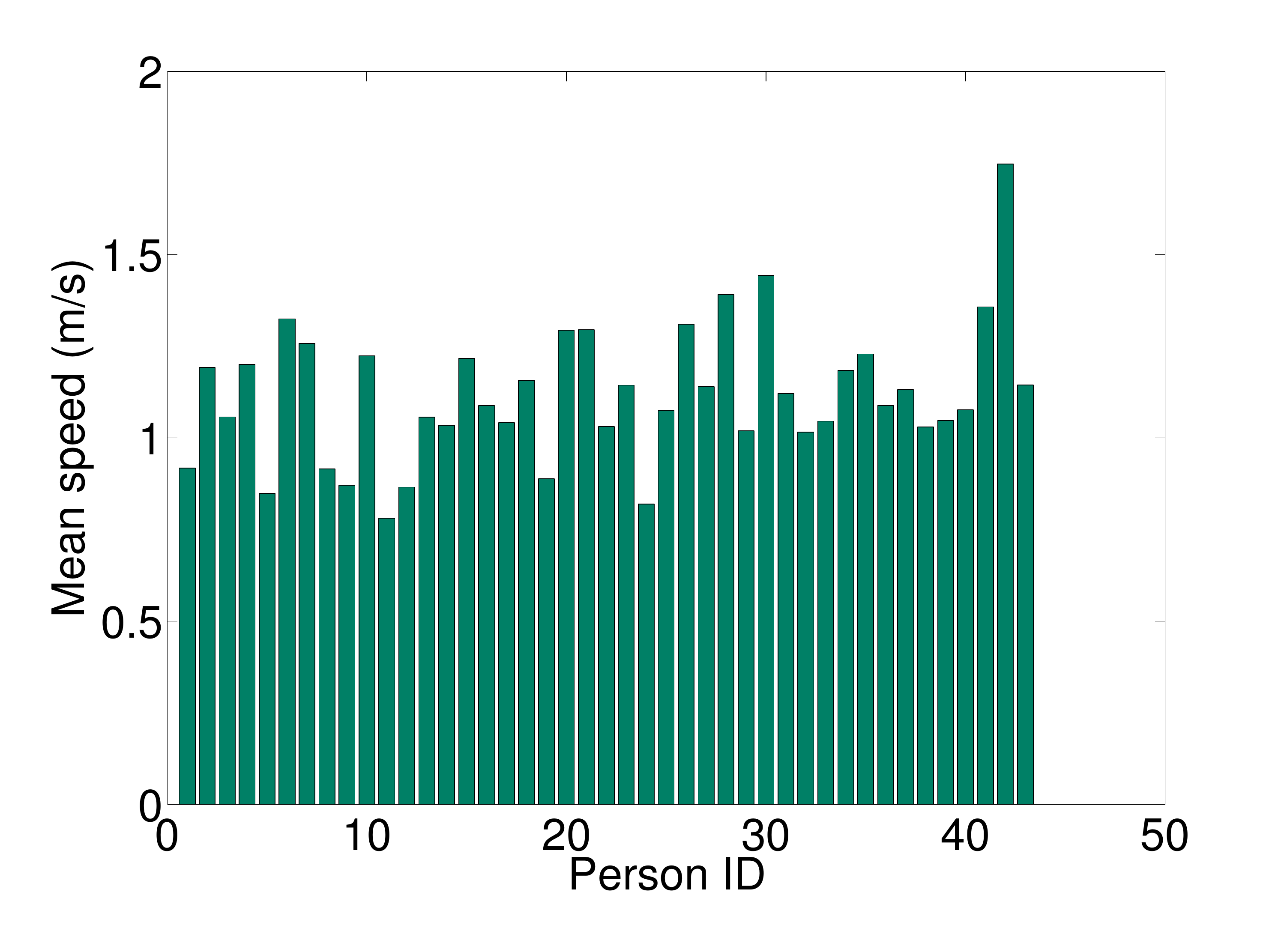} 
\label{fig6s}}
\subfigure[AVG-TownCentre]{
\includegraphics[trim=2.0cm 0.5cm 2.2cm 0.9cm,clip=true,width=0.22\linewidth]{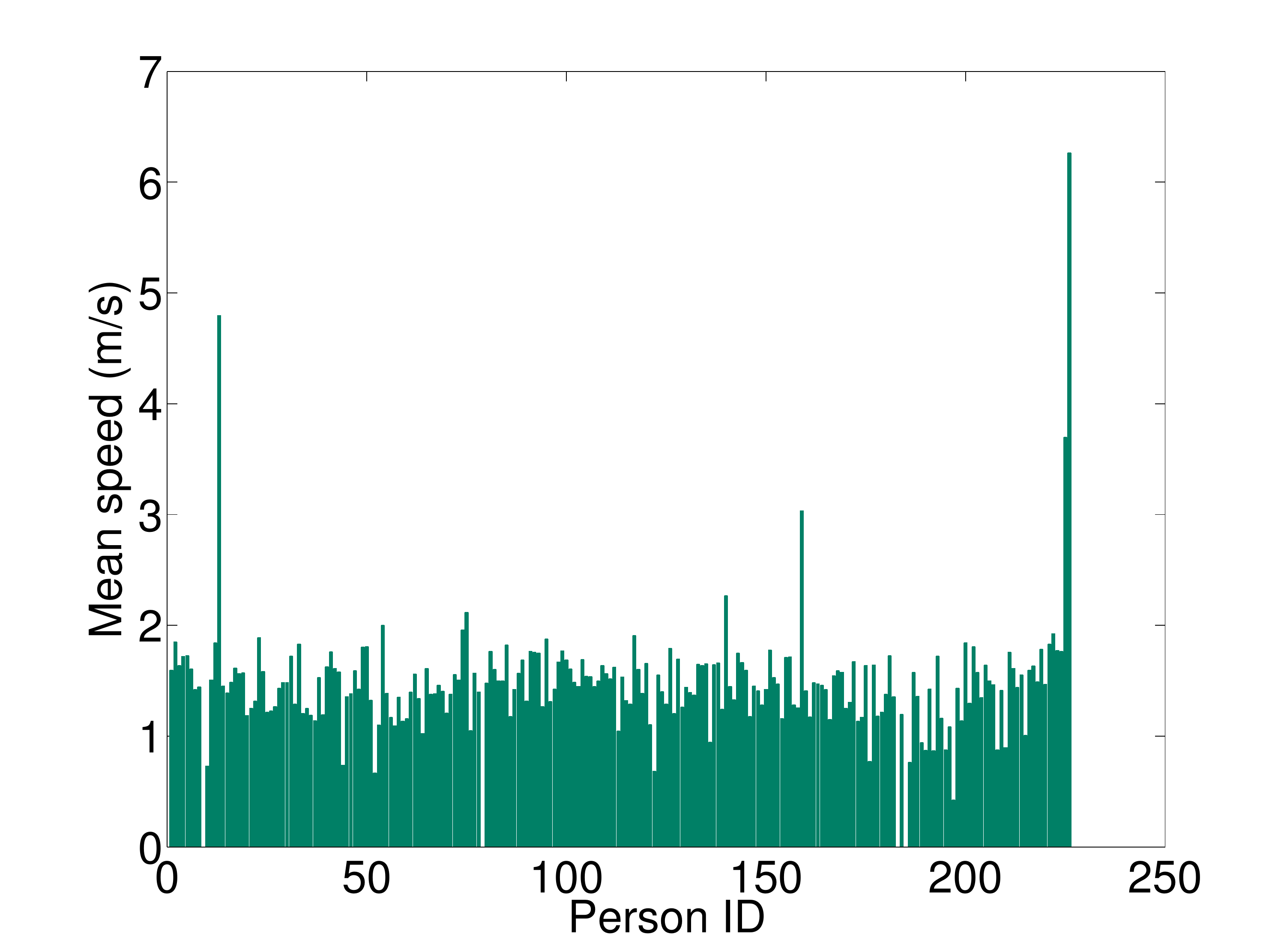} 
\label{fig7s}}
\subfigure[TUD-Stadtmitte]{
\includegraphics[trim=1.0cm 0.5cm 2.2cm 0.9cm,clip=true,width=0.23\linewidth]{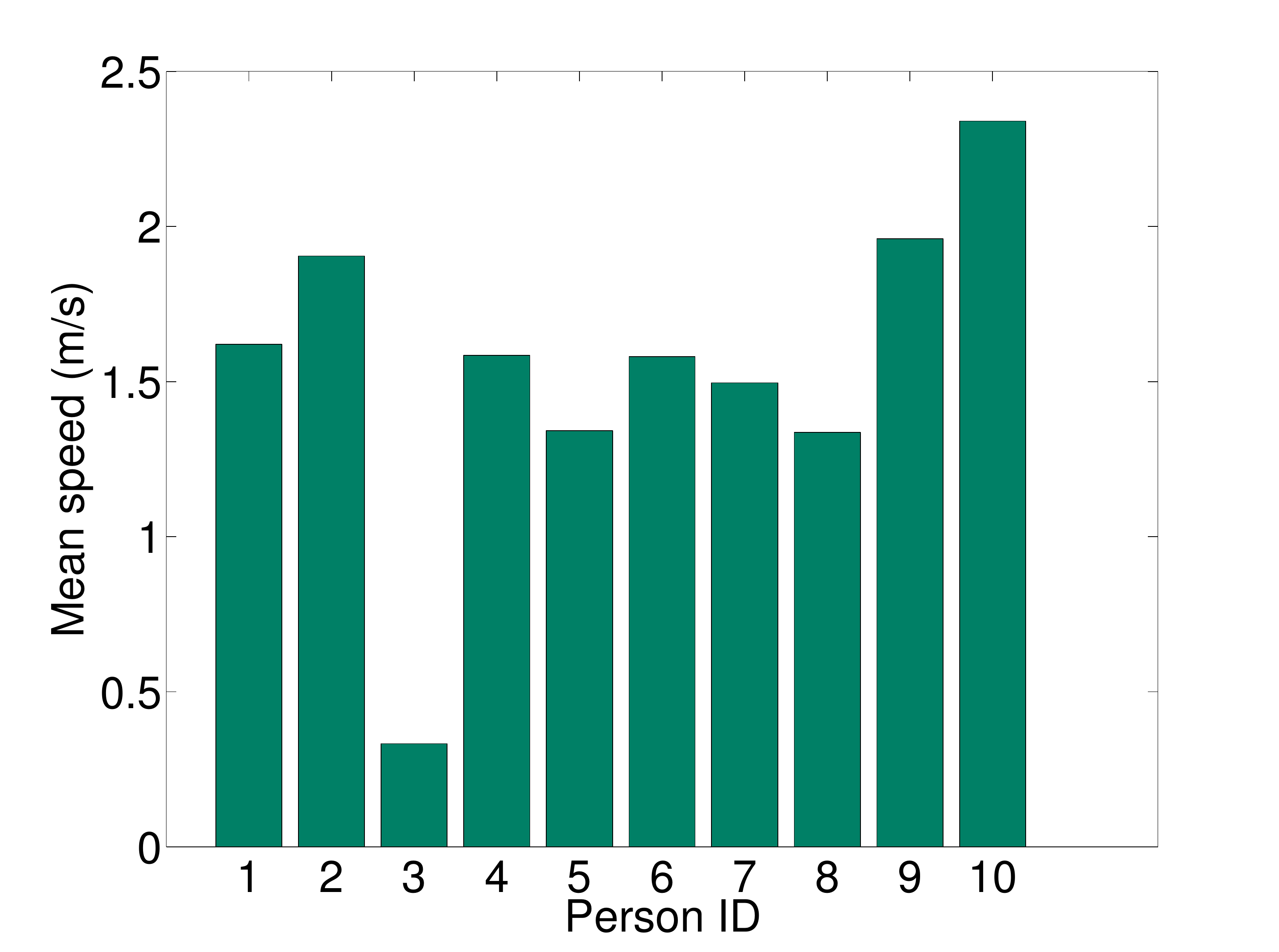} 
\label{fig8s}}
\caption{{\it Top row:} Pedestrian speed histograms per sequence. {\it Bottom row:} Mean speed per pedestrian per sequence.}
\label{fig:3Ddatastatic}
\end{figure*}

For the 4 sequences filmed using a static camera, AVG-TownCentre, PETS09-S2L1, PETS09-S2L2 and TUD-Stadtmitte, the calibration files from the sources \cite{Benfold:2011:CVPR, Andriluka:2010:CVPR, Ferryman:2010:PETS} are used to compute a 2D homography between the image plane and the ground plane. All $z$ coordinates are set to $0$, indicating the position of the feet of the pedestrian.
In order to measure the accuracy of the calibration for each sequence,
we use the manually annotated trajectories and plot the velocities of
the pedestrians at each frame. Realistic walking speeds range from 0
-- 3~$\text{m}/\text{s}$, with a mean comfortable walking speed of
1.4~$\text{m}/\text{s}$. This is confirmed by the distribution that we
see in \Fig~\ref{fig1s}. The other sequences
(Figs.~\ref{fig2s}-\ref{fig4s}) have a few speeds in the range of 3
-- 10~$\text{m}/\text{s}$. These are not real speeds, since there are
no running pedestrians in the sequences. These outliers
are largely due to projective geometry.
For example, variations in the size of the bounding box can introduce artificial shifts in 2D that greatly affect the 3D position in the scene. 

The bottom row of \Fig~\ref{fig:3Ddatastatic} shows the mean
speeds per pedestrian and sequence.
As we can see, most pedestrians walk at a speed between 1 --
1.5~$\text{m}/\text{s}$, hence we can conclude that the calibration is
accurate enough for tracking.

\begin{figure*}[htpb]
\centering
\includegraphics[width=\linewidth]{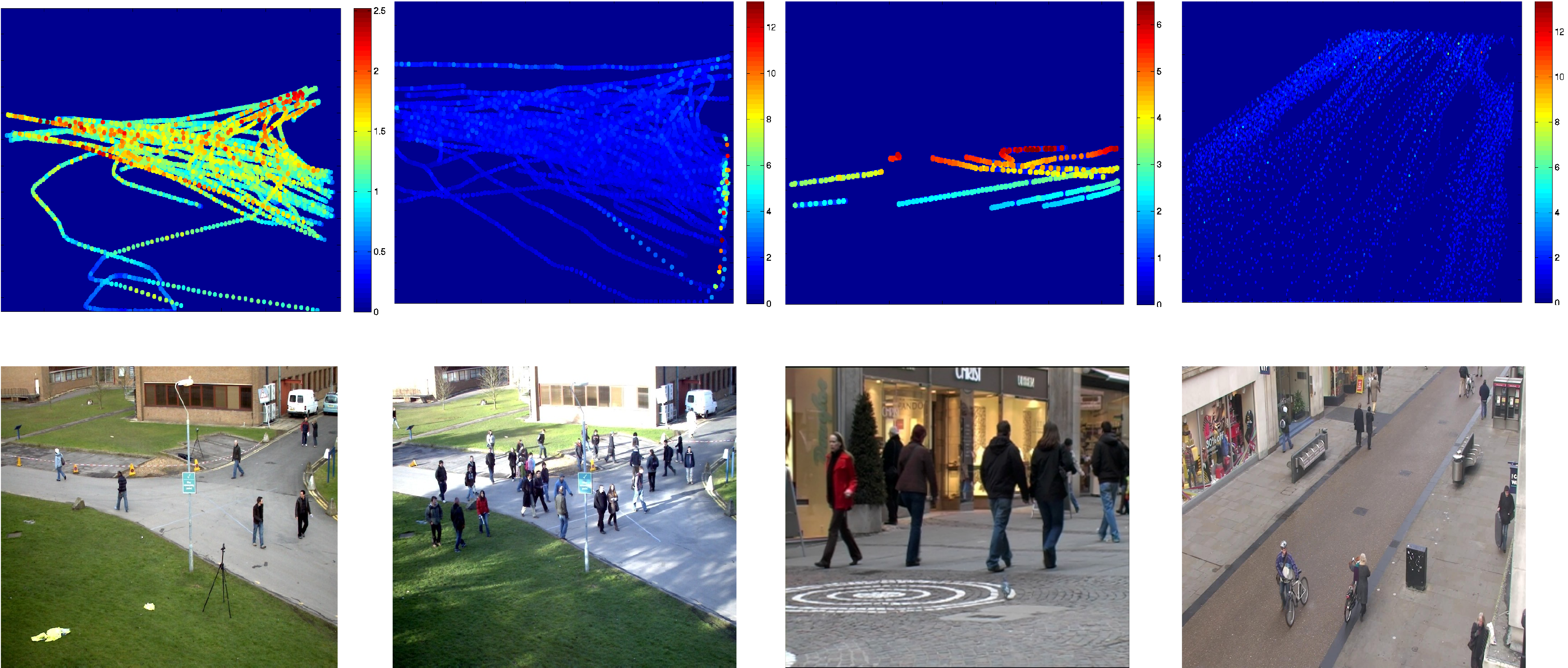} 
\caption{{\it Top row:} Speed distributions in image space for the PETS09-S2L1, PETS09-S2L2, AVG-TownCentre and TUD-Stadtmitte sequences, respectively. Note that the scale is different for each image. {\it Bottom row:} Sample frame for each sequence.}
\label{fig:3Dmapsstatic}
\end{figure*}

We can further analyze the spurious high speeds that we observe in
some sequences by plotting a speed distribution in image space as shown
in the top row of \Fig \ref{fig:3Dmapsstatic}. For PETS09-S2L2, we can
see that most artifacts are concentrated on the bottom-right part of
the image, where pedestrians leave the scene.
The fact that a leaving pedestrian is cropped by the image border
makes the bounding box around it thinner (following the annotation
policy of PETS). Since the bottom-center of the bounding box is the 
2D position used to obtain the 3D information, the 2D position is
likely shifted away from the real position of the pedestrian's feet. 
In the case of AVG-TownCentre, we can see some points on the image
where unusually high speeds are observed. These are typically far away
from the camera and present at the beginning or the end of the
sequence, where correct annotation is difficult. This also accounts
for the peaks of mean speed in \Fig \ref{fig7m}, which belong to two
pedestrians observed for 2 frames and are far away from the
camera. Their position fluctuation is a simple artifact of variability
in the bounding box placement. 
Finally, for the TUD-Stadtmitte sequence, we clearly see that high velocities are concentrated in the part of the image that is far away from the camera, and therefore also far away from the points used for calibration. The bounding box shifts in that area have a bigger impact on the 3D position. 
Avoiding the effects discussed would require new ``3D aware''
annotations for each sequence, which we leave for future work.

\subsubsection{Moving camera sequences}

For the sequences with moving cameras, the authors \cite{Ess:2008:CVPR} provide one file for each image, containing the calibration of the left camera, which allows us to backproject the feet of the person to the world coordinate system and to find the 3D position by intersecting the ray with the estimated ground plane.
The error of this calibration increases significantly as pedestrians move away from the camera, which makes tracking of far away objects very imprecise.

\begin{figure*}[htpb]
\centering
\subfigure[ETH-Bahnhof]{
\includegraphics[trim=2.0cm 0.5cm 2.2cm 0.9cm,clip=true,width=0.23\linewidth]{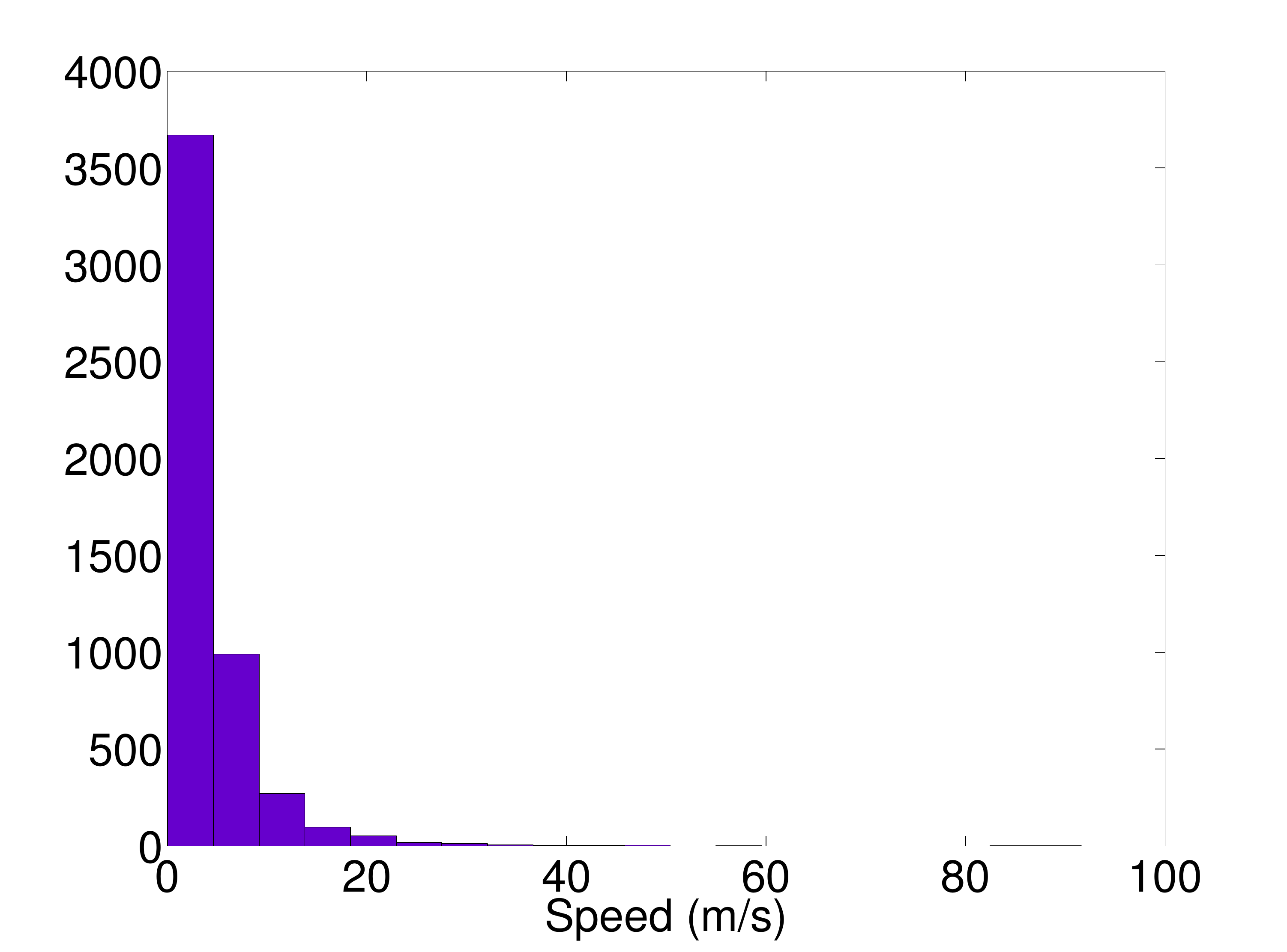} 
\label{fig1m}}
\subfigure[ETH-Sunnyday]{
\includegraphics[trim=2.0cm 0.5cm 2.2cm 0.9cm,clip=true,width=0.23\linewidth]{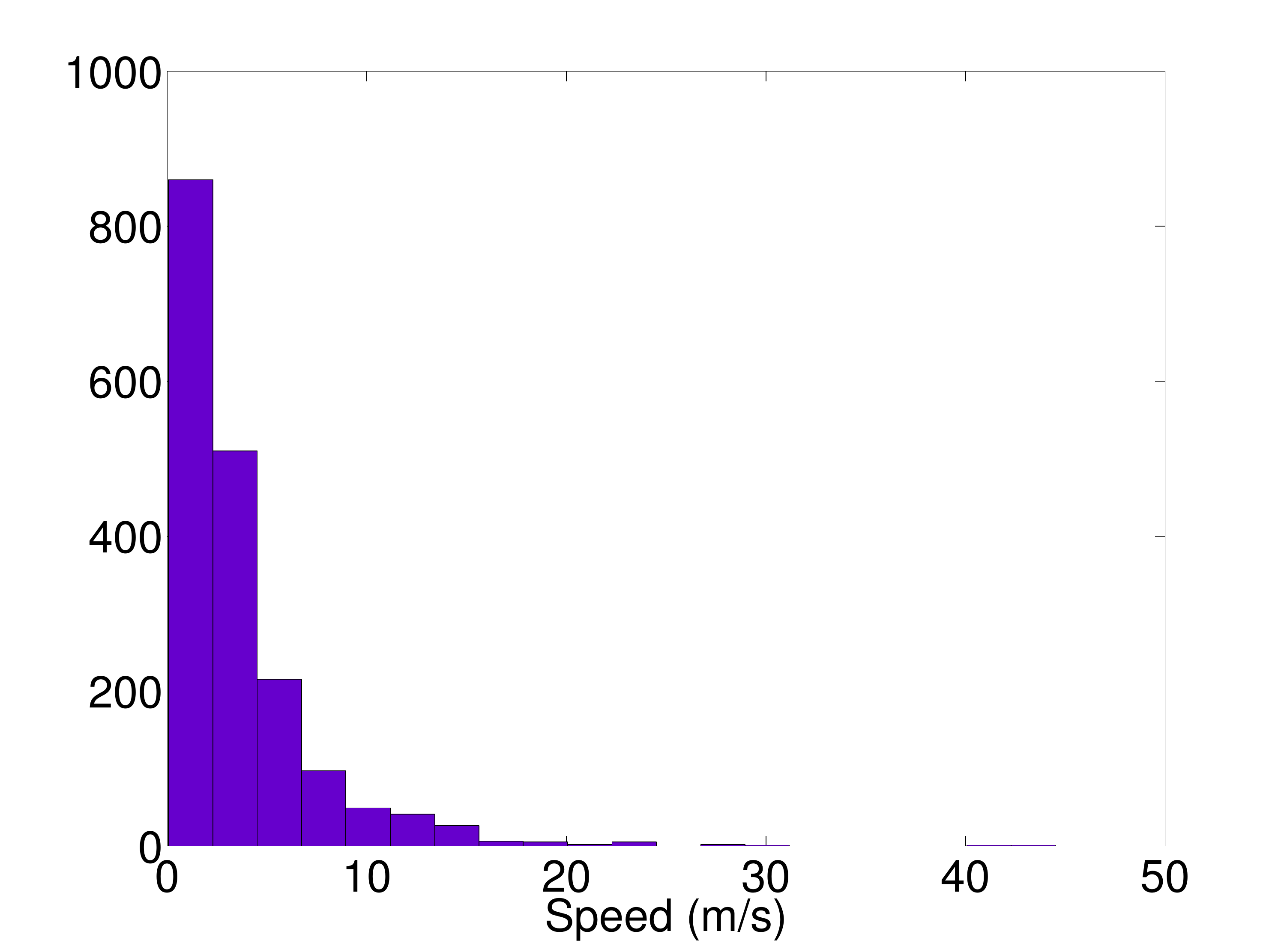} 
\label{fig2m}}
\subfigure[ETH-Linthescher]{
\includegraphics[trim=2.0cm 0.5cm 2.2cm 0.9cm,clip=tru ,width=0.23\linewidth]{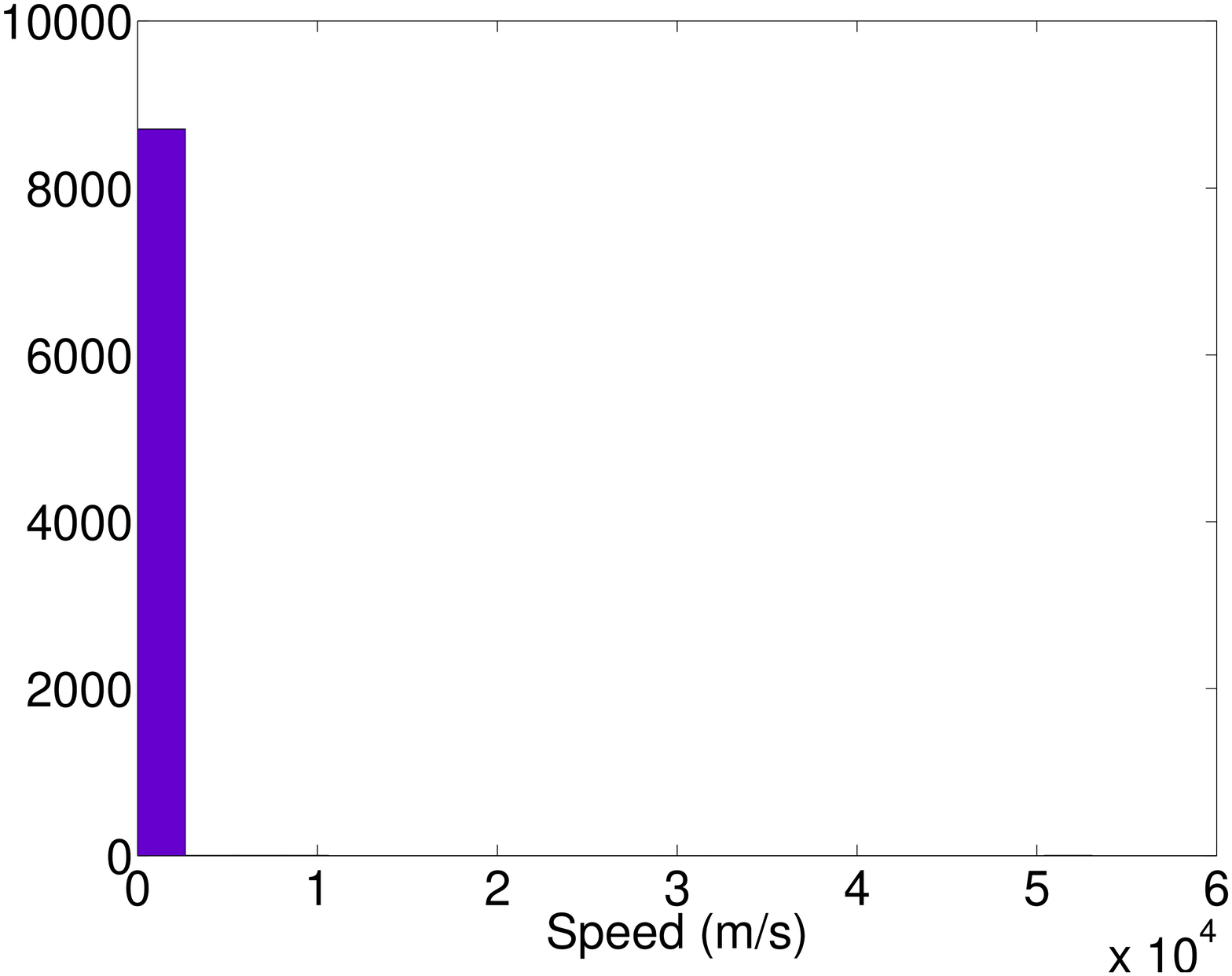} 
\label{fig3m}}
\subfigure[ETH-Jelmoli]{
\includegraphics[trim=2.0cm 0.5cm 2.2cm 0.9cm,clip=true,width=0.23\linewidth]{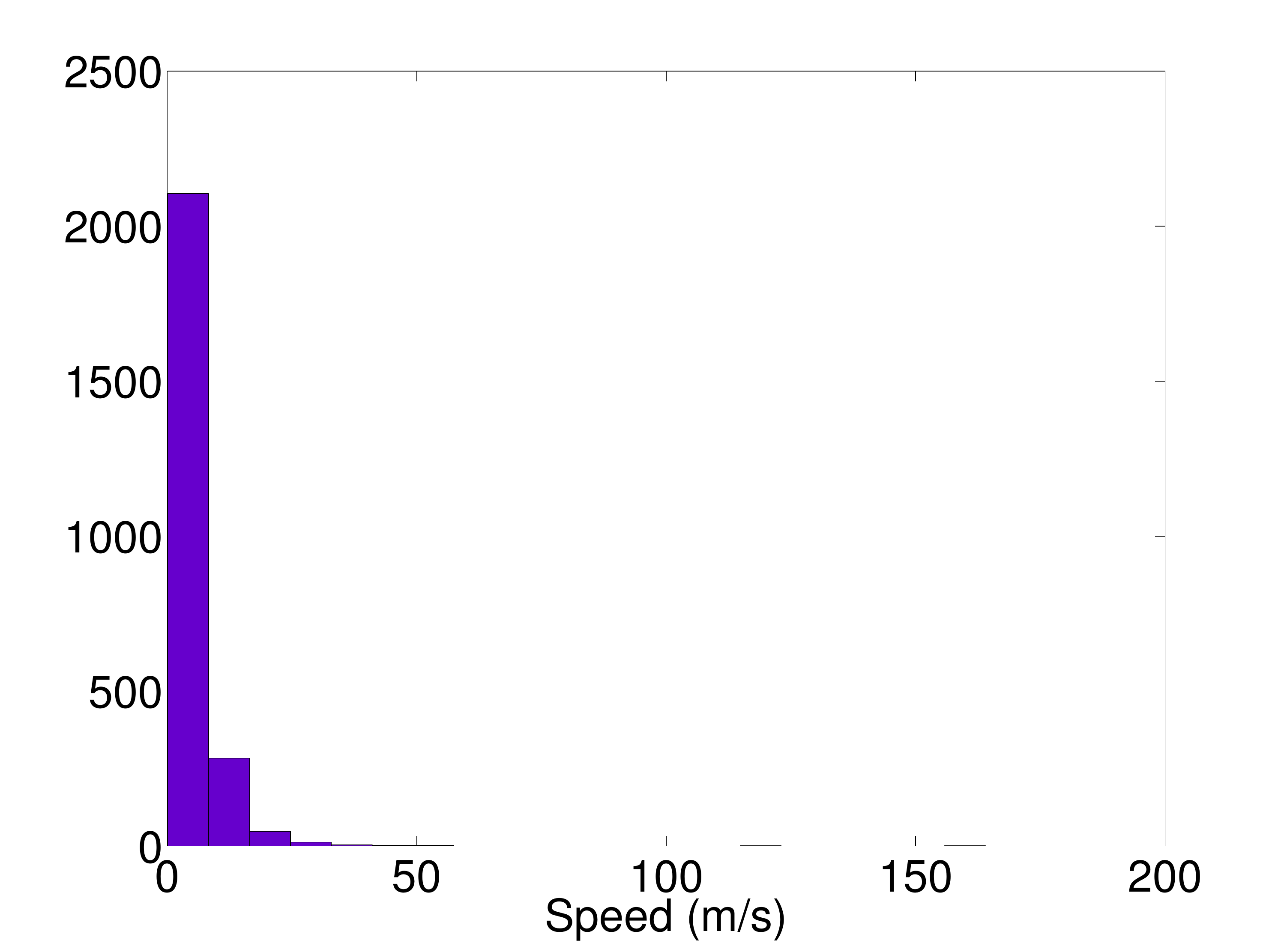} 
\label{fig4m}}
\subfigure[ETH-Bahnhof]{
\includegraphics[trim=1.0cm 0.5cm 2.2cm 0.9cm,clip=true,width=0.23\linewidth]{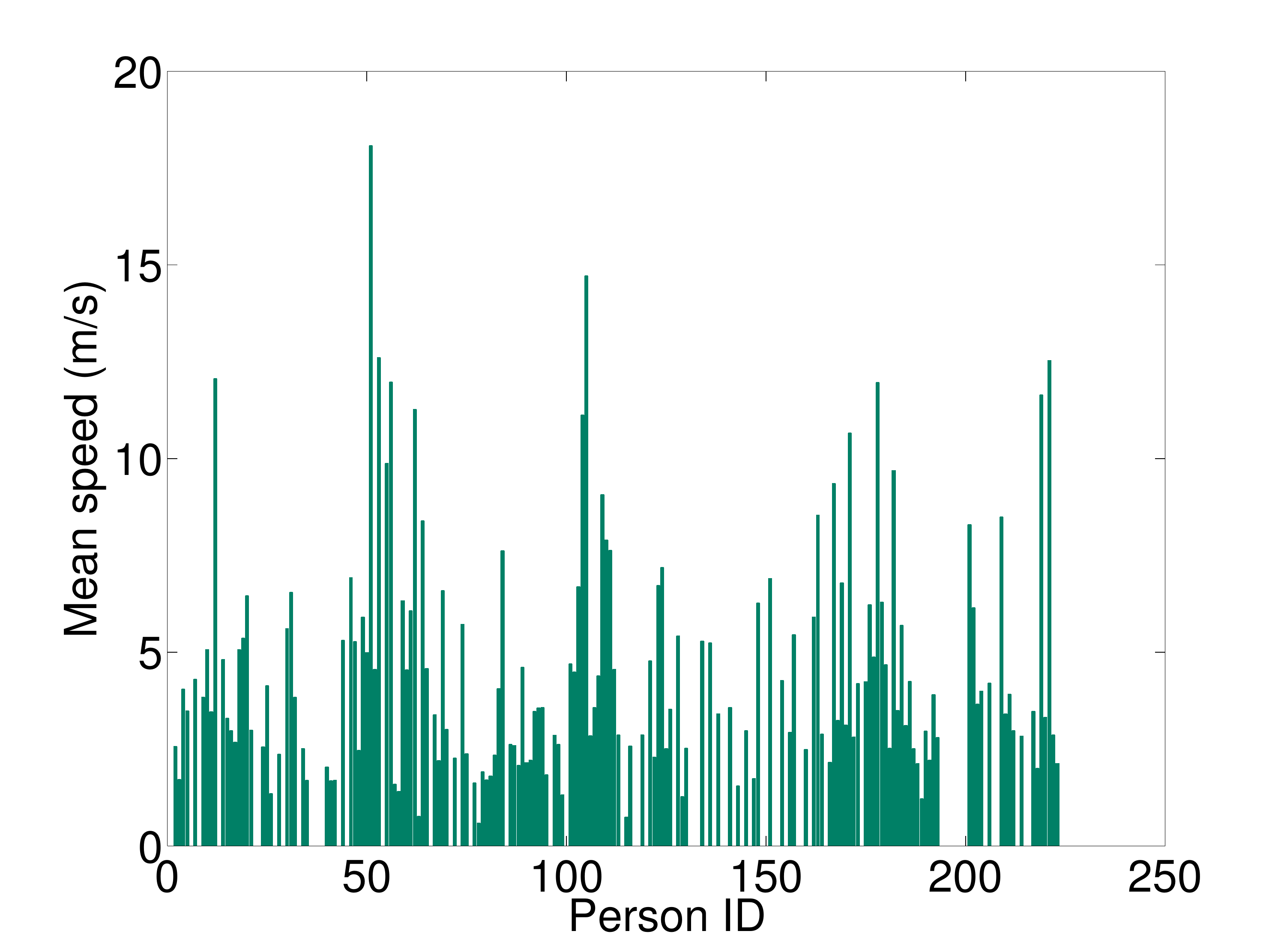} 
\label{fig5m}}
\subfigure[ETH-Sunnyday]{
\includegraphics[trim=1.0cm 0.5cm 2.2cm 0.9cm,clip=true,width=0.23\linewidth]{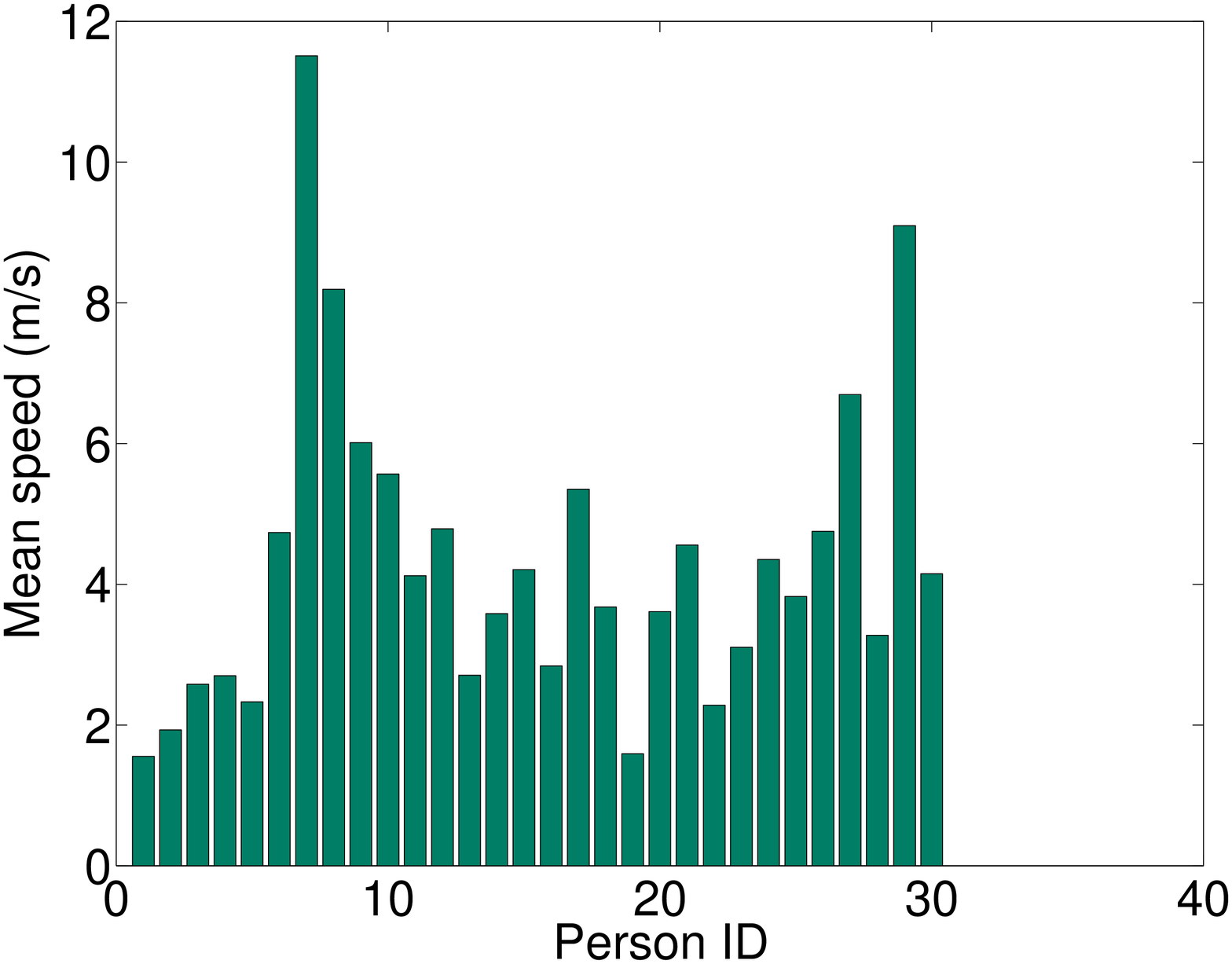} 
\label{fig6m}}
\subfigure[ETH-Linthescher]{
\includegraphics[trim=1.0cm 0.5cm 2.2cm 0.9cm,clip=true,width=0.23\linewidth]{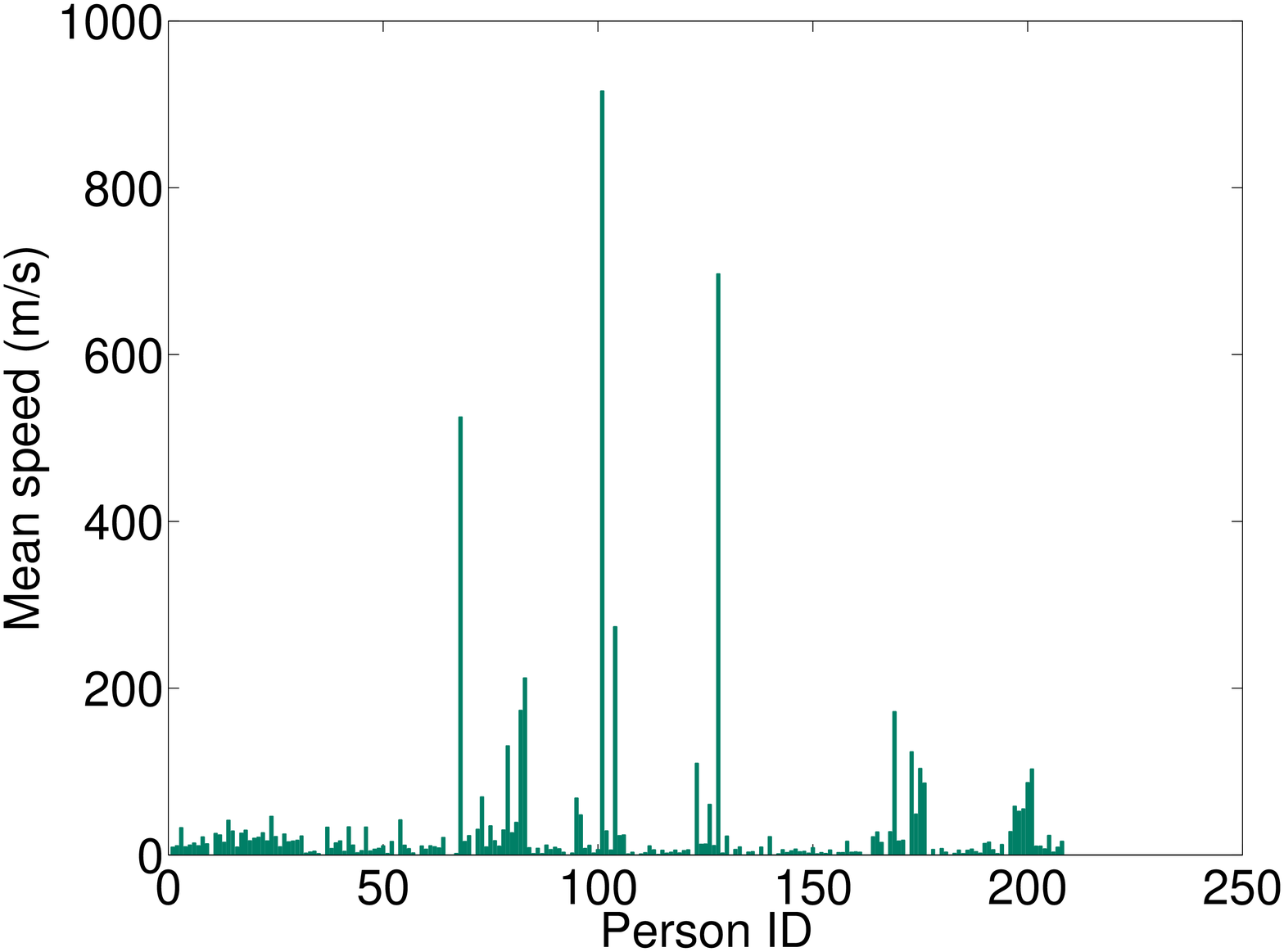} 
\label{fig7m}}
\subfigure[ETH-Jelmoli]{
\includegraphics[trim=1.0cm 0.5cm 2.2cm 0.9cm,clip=true,width=0.23\linewidth]{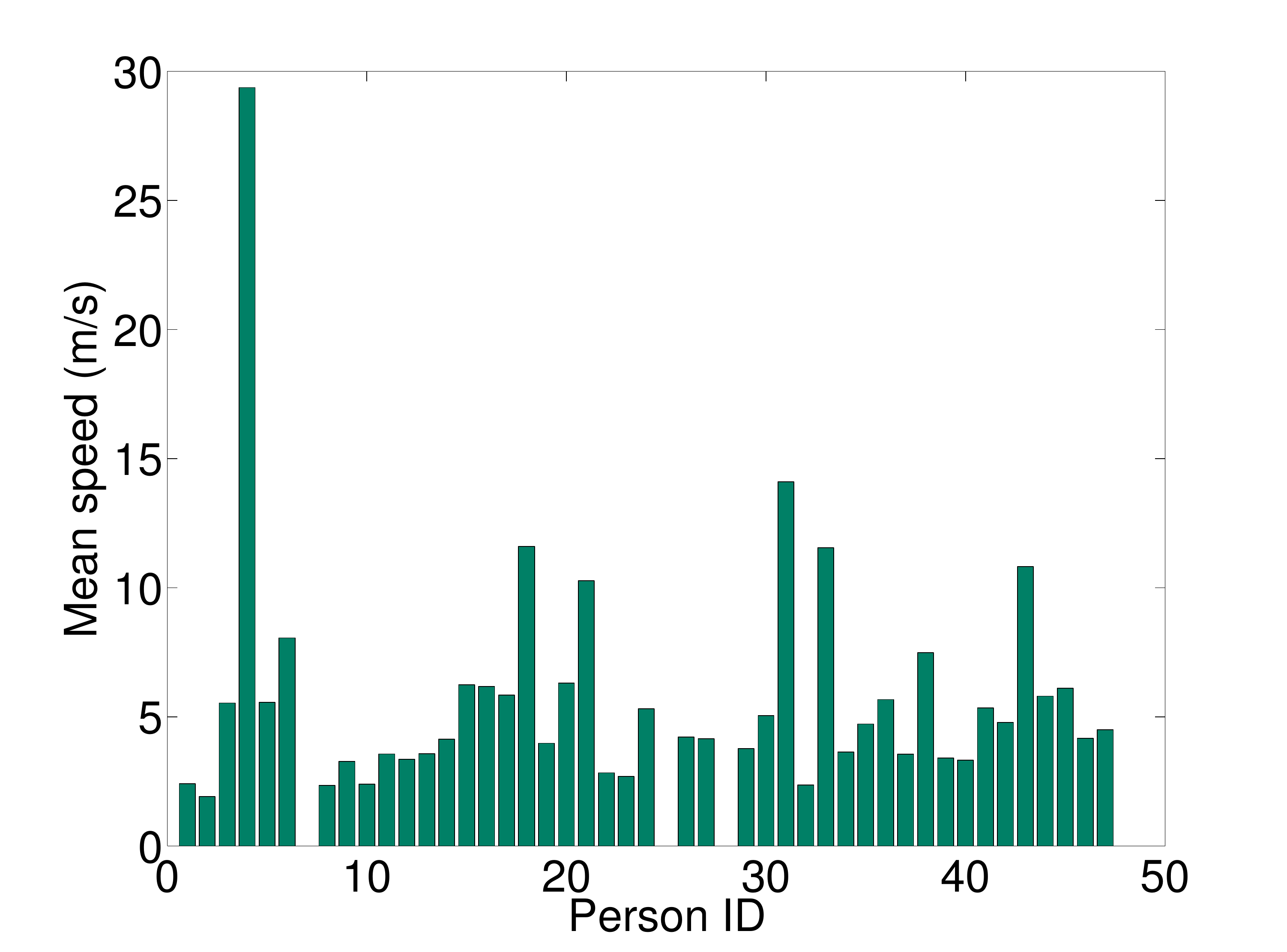} 
\label{fig8m}}
\caption{{\it Top row:} Pedestrian speed histograms per sequence. {\it Bottom row:} Mean speed per pedestrian per sequence.}
\label{fig:3Ddatamoving}
\end{figure*}

\begin{figure*}[htpb]
 \centering
 \includegraphics[width=\linewidth]{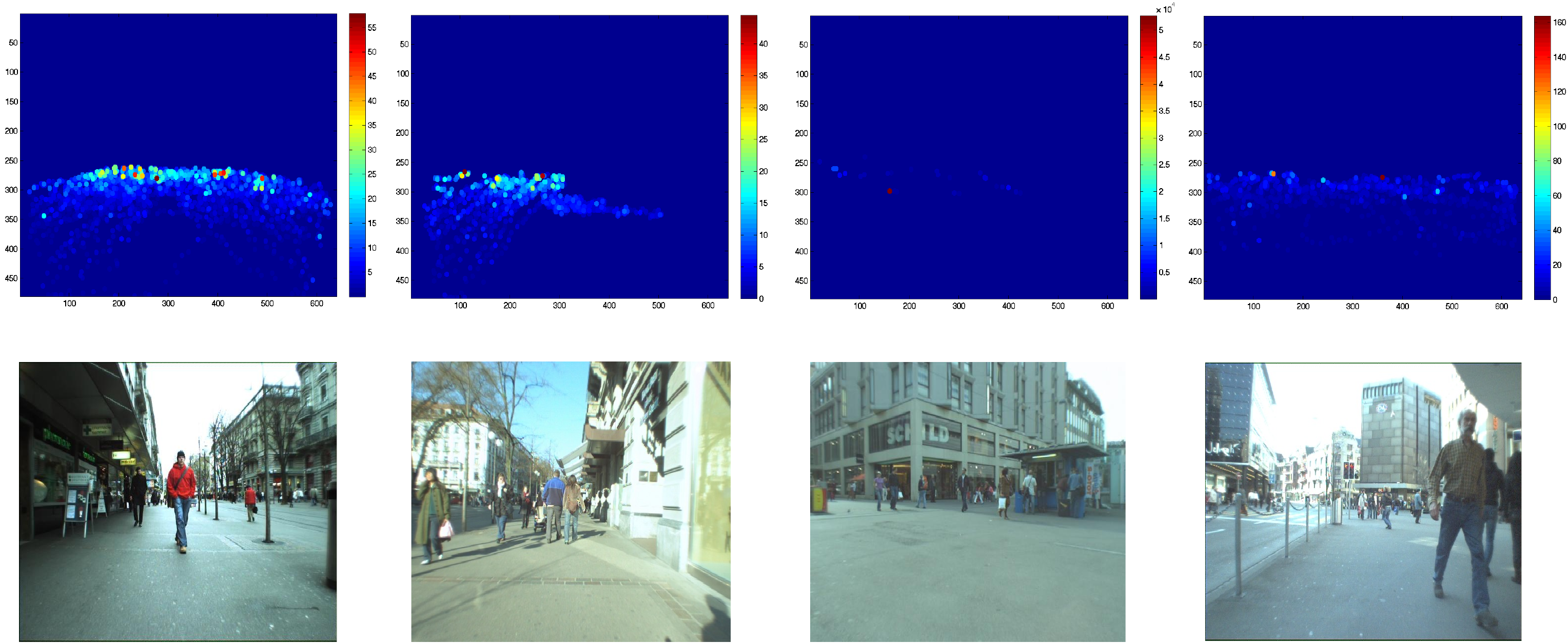} 
 \caption{{\it Top row:} Speed distributions in image space for the ETH-Bahnhof, ETH-Sunnyday, ETH-Linthescher and ETH-Jelmoli sequences, respectively. Note that the scale is different for each image. {\it Bottom row:} Sample frame for each sequence.}
 \label{fig:3Dmapsmoving}
 \end{figure*}

We do the same velocity analysis for these sequences, shown in \Fig
\ref{fig:3Ddatamoving}, and observe that some of the velocities reach
200 -- 700~$\text{m}/\text{s}$, indicating a clear problem in the 3D
position estimation. Looking at the mean velocity per pedestrian, we
can observe very high velocities, especially for ETH-Bahnhof and
ETH-Linthescher sequences. These are mostly pedestrians walking far
away from the camera and usually visible for a short period of
time. This is further observable in the maps of \Fig
\ref{fig:3Dmapsmoving}, where we can see that these incorrect peak
velocities are found mostly in the region far away ($\sim 10$ m) from
the camera.
This again illustrates the challenge of obtaining accurate 3D
information from 2D bounding boxes, simply due to the nature of
projective geometry.
For these sequences, the added inaccuracy introduced by the  
automatic ground plane angle estimation makes the 3D information unreliable, which is why we decided not to include these sequences in the 3D benchmark. 

As future work, we plan on using the additional view provided for these
sequences to strengthen the 3D estimation.
Ideally and for all sequences, the pedestrian's feet should be
annotated directly, since in general annotations for 2D and 3D
tracking purposes may differ.
Further annotation issues are discussed in \Sec \ref{sec:annotations}.

\subsection{Detections}
\label{sec:detections}
\begin{figure*}[t]
\centering
\subfigure[][Detection performance of \cite{Dollar:2014:PAMI}]{
\includegraphics[width=4cm,height=3cm]{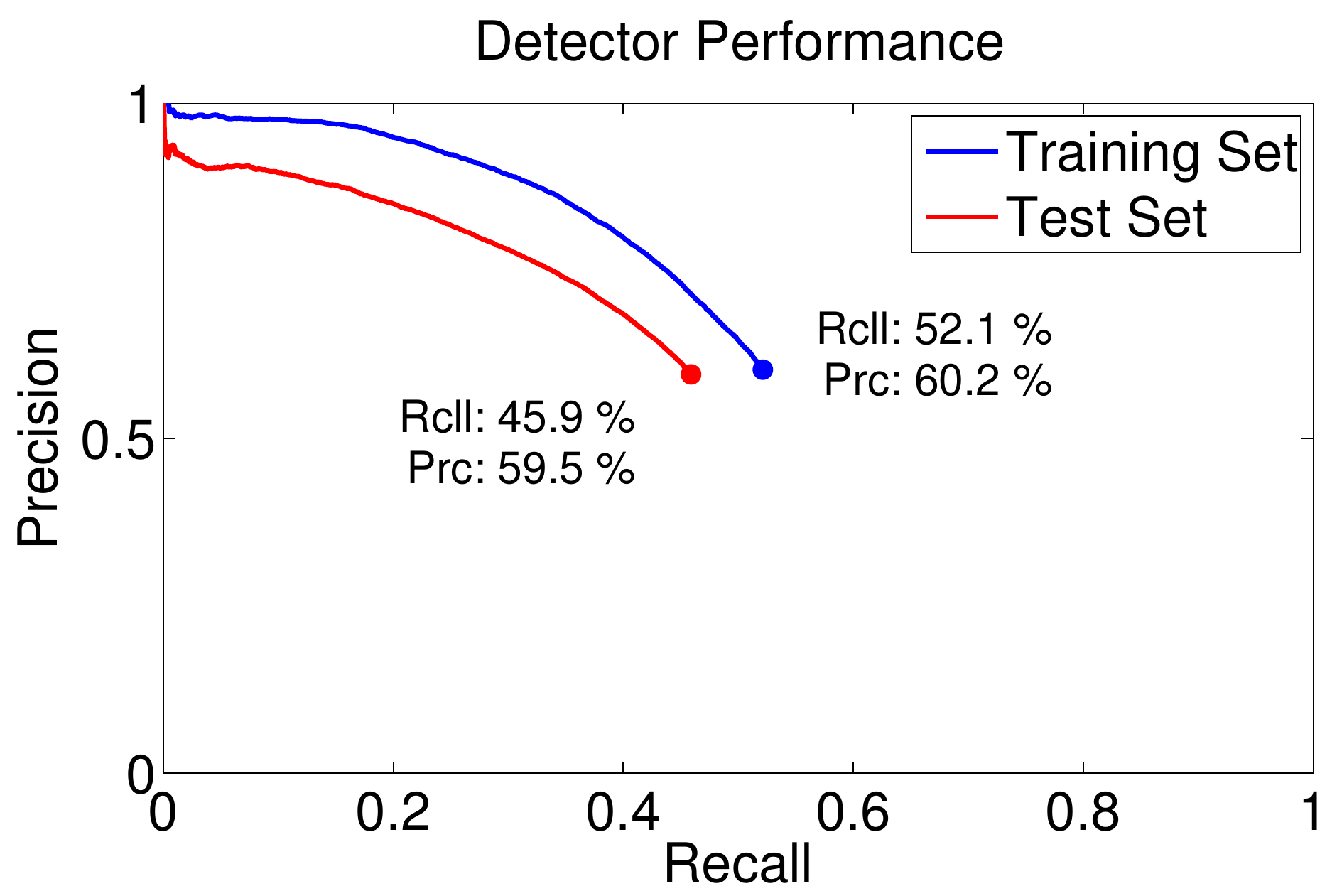}} 
\hspace{1cm}
\subfigure[][ADL-Rundle-8]{
\includegraphics[width=4cm,height=3cm]{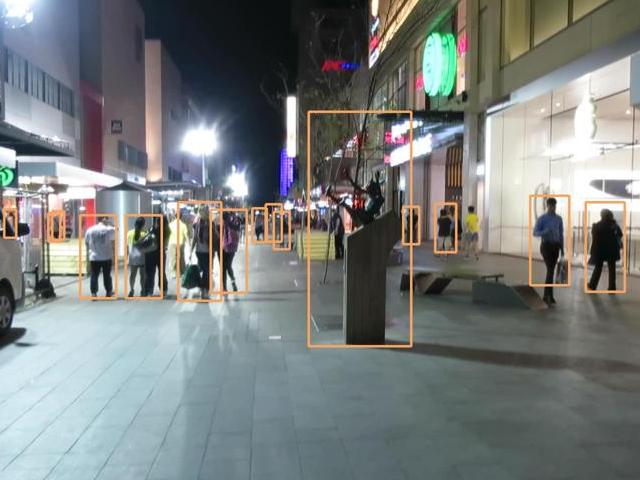}} 
\subfigure[][Venice-1]{
\includegraphics[width=4cm,height=3cm]{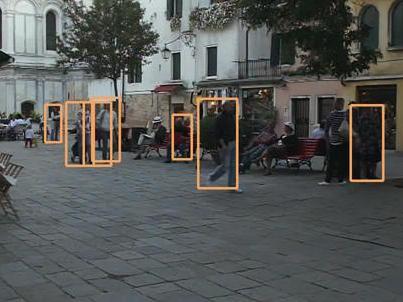}} 
\subfigure[][KITTI-16]{
\includegraphics[width=4cm,height=3cm]{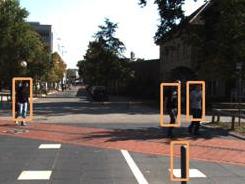}} 
\caption{(a) The performance of the provided detection bounding boxes
evaluated on the training (blue) and the test (red) set. The circle indicates
the operating point (\ie the input detection set) for the trackers. (b-d) Exemplar detection results.}
\label{fig:det-performance}
\end{figure*}
To detect pedestrians in all images, 
we use the recent object detector implementation of Doll\'{a}r \etal 
\cite{Dollar:2014:PAMI}, which is based on aggregated channel features 
(ACF). We rely on the default parameters and the pedestrian model 
trained on the INRIA dataset \cite{Dalal:2005:CVPR}, rescaled with a
factor of $0.6$ to enable the detection of smaller pedestrians. The minimal 
bounding box height in our benchmark is 59 pixels. The detector performance
along with three sample frames is depicted in \Fig~\ref{fig:det-performance} for both the training and 
the test set of the benchmark. Note that the recall does not reach 100\% 
because of the non-maximum suppression applied.

Obviously, we cannot (nor necessarily want to) prevent anyone from
using a different set of detections, or even rely on a different set
of features altogether.
However, we require that this is noted as part of the tracker's 
description and is also displayed in the ratings table.

\subsection{Data format}
\label{sec:data-format}
All images were converted to JPEG and named sequentially to a 6-digit file name (\eg~000001.jpg). Detection and annotation files are simple comma-separated value (CSV) files. Each line represents one object instance, and it contains 10 values as shown in \Tab \ref{tab:dataformat}.

The first number indicates in which frame the object appears, while
the second number identifies that object as belonging to a trajectory
by assigning a unique ID (set to $-1$ in a detection file, as no ID is
assigned yet). Each object can be assigned to only one trajectory.
The next four numbers indicate the position of the bounding box of the
pedestrian in 2D image coordinates. The position is indicated by the
top-left corner as well as width and height of the bounding box.
This is followed by a single number, which in case of detections
denotes their confidence score.
The last three numbers indicate the 3D position in real-world coordinates of the pedestrian. This position represents the feet of the person. In the case of 2D tracking, these values will be ignored and can be left at $-1$.

\begin{table*}[hbt]
\begin {center}
 \begin{tabular}{| c | c| p{13cm}|}
 \hline
     \bf Position & \bf Name & \bf Description\\ 
          \hline 
    1 & Frame number & Indicate at which frame the object is present\\
 2 & Identity number & Each pedestrian trajectory is identified by a
 unique ID ($-1$ for detections)\\
 3 & Bounding box left &  Coordinate of the top-left corner of the pedestrian bounding box\\
  4 & Bounding box top & Coordinate of the top-left corner of the pedestrian bounding box \\
 5 & Bounding box width & Width in pixels of the pedestrian bounding box\\
 6 & Bounding box height & Height in pixels of the pedestrian bounding box\\
 7 & Confidence score & Indicates how confident the detector is that this instance is a pedestrian. 
 For the ground truth and results, it acts as a flag whether the entry is to be considered.  \\
  8 &  $x$ & 3D $x$ position of the pedestrian in real-world
  coordinates ($-1$ if not available)\\
 9 & $y$ &  3D $y$ position of the pedestrian in real-world coordinates ($-1$ if not available)\\
 10 & $z$ & 3D $z$ position of the pedestrian in real-world
 coordinates  ($-1$ if not available)\\
      \hline 
    \end{tabular}
  \end{center}
    \caption{Data format for the input and output files, both for detection and annotation files.}
\label{tab:dataformat}
\end{table*}

 An example of such a detection 2D file is:
 
 \begin{samepage}
\begin{center}
\begin{footnotesize}
  \texttt{1, -1, 794.2, 47.5, 71.2, 174.8, 67.5, -1, -1, -1}\nopagebreak\\
  \texttt{1, -1, 164.1, 19.6, 66.5, 163.2, 29.4, -1, -1, -1}\nopagebreak\\
  \texttt{1, -1, 875.4, 39.9, 25.3, 145.0, 19.6, -1, -1, -1}\nopagebreak\\
  \texttt{2, -1, 781.7, 25.1, 69.2, 170.2, 58.1, -1, -1, -1}\nopagebreak\\
\end{footnotesize}
\end{center}
\end{samepage}

For the ground truth and results files, the 7$^\text{th}$ value (confidence score) acts as a flag whether the entry is to be considered. A value of 0 means that this particular instance is ignored in the evaluation, while a value of 1 is used to mark it as active. 

 An example of such an annotation 2D file is:
 
 \begin{samepage}
\begin{center}
\begin{footnotesize}
  \texttt{1, 1, 794.2, 47.5, 71.2, 174.8, 1, -1, -1, -1}\nopagebreak\\
  \texttt{1, 2, 164.1, 19.6, 66.5, 163.2, 1, -1, -1, -1}\nopagebreak\\
  \texttt{1, 3, 875.4, 39.9, 25.3, \enspace35.0, 0, -1, -1, -1}\nopagebreak\\
  \texttt{2, 1, 781.7, 25.1, 69.2, 170.2, 1, -1, -1, -1}\nopagebreak\\
\end{footnotesize}
\end{center}
\end{samepage}

In this case, there are 2 pedestrians in the first frame of the sequence, with identity tags 1, 2. The third pedestrian is too small and therefore not considered, which is indicated with a flag value (7$^\text{th}$ value) of 0. In the second frame, we can see that pedestrian 1 remains in the scene. Note, that since this is a 2D annotation file, the 3D positions of the pedestrians are ignored and therefore are set to -1. 
Note that all values including the bounding box are 1-based, \ie the top left corner corresponds to $(1,1)$.

To obtain a valid result for the entire benchmark, a separate CSV file following the format described above must be created for each sequence and called \texttt{``Sequence-Name.txt''}. All files must be compressed into a single zip file that can then be uploaded to be evaluated.

\subsection{Expansion through crowdsourcing}

We foresee a yearly update of the benchmark datasets in order to include new, more challenging sequences and eventually remove outdated or repetitive sequences. The goal is to push forward research in multi-target tracking by increasing the difficulty of the data as new, more accurate methods are proposed by the community. 
We want to make a call to the community to share their sequences, detections or annotations to the benchmark, so as to include a large variety of data. More importantly, the goal is to increase the type of data to the following categories:\\

\begin{itemize}
 \item Tracking of cars, bicycles, etc.~in outdoor scenarios;
 \item Biological data such as cell, bird or fish tracking;
 \item Sports data: basketball games, hockey or soccer;
 \item Large-scale multi-view sequences.\\
\end{itemize}

Sequences of such scenarios do exist in the literature, \emph{e.g.}~thousands of bats filmed using thermal cameras \cite{wuiccv2009}, cell tracking data \cite{limia2008}, basketball game \cite{berclaztpami2011}, hockey game \cite{okumaeccv2004}, several indoor multi-view sequences \cite{berclaztpami2011}, or recent large-scale multi-view sequences containing millions of pedestrian trajectories \cite{alahicvpr2014}.
We encourage the community to contact us with any interesting data that they would like included in the benchmark structure, and we commit to extending the benchmark to other interesting and relevant categories. Each category will have its own separate submissions.

\section{Baseline Methods}
\label{sec:baselines}
As a starting point for the benchmark, we have included a number of 
recent multi-target tracking approaches as baselines, which we will 
briefly outline for completeness but refer the reader to the respective 
publication for more details. Note that we have used the publicly 
available code and trained all of them in the same way\footnote{Except 
for \Tracker{TBD}, which does not disclose any obvious free parameters.}
(\cf~\Sec~\ref{sec:training}). 
However, we explicitly state that the 
provided numbers may not represent the best possible performance for 
each method, as could be achieved by the authors themselves. 
Table~\ref{tab:results} lists current benchmark results for all 
baselines as well as for all anonymous entries at the time of 
writing of this manuscript.

\subsection{Training and testing}
\label{sec:training}
Most of the available tracking approaches do not include a learning (or 
training) algorithm to determine the set of model parameters for a 
particular dataset. Therefore, we follow a simplistic search scheme for 
all baseline methods to find a good setting for our benchmark. To that 
end, we take the default parameter set $\parvec := 
\{\onepar_1,\ldots,\onepar_P\}$ as suggested by the authors, where $P$ is 
the number of free parameters for each method. We then perform 100 
independent runs 
on the training set with varying parameters. In each run, a parameter 
value $\onepar_i$ is uniformly sampled around its default value in the 
range $[\frac{1}{2}\onepar_i , 2 \onepar_i]$. Finally, the parameter set 
$\parvec^*$ that achieved the highest MOTA score across all 100 runs
(\cf~\Sec~\ref{sec:mota}) is taken as the optimal setting and run once 
on the test set. The optimal parameter set is stated in the description
entry for each baseline method on the benchmark website.

\subsection{\Tracker{DP\_NMS}: Network flow tracking}
\label{sec:DP_NMS}
Since its original publication \cite{Zhang:2008:CVPR}, a large number of 
methods that are based on the network flow formulation have appeared in 
the literature \cite{Pirsiavash:2011:CVPR, Butt:2013:CVPR, 
Liu:2013:CVPR, Wang:2014:CVPR, lealcvpr2012}. The basic idea is to model the tracking 
as a graph, where each node represents a detection and each edge 
represents a transition between two detections. Special source and sink 
nodes allow spawning and absorbing trajectories. A solution is obtained 
by finding the minimum cost flow in the graph. Multiple assignments and track 
splitting is prevented by introducing binary and linear constraints.

Here we use two solvers: (i) the successive shortest paths approach 
\cite{Pirsiavash:2011:CVPR} that employs dynamic programming with
non-maxima suppression, termed \Tracker{DP\_NMS}; (ii) a linear programming solver 
that we use for both 2D and 3D data (\Tracker{LP2D} and \Tracker{LP3D}, respectively),
and that appears as a baseline in \cite{lealcvpr2014}.
This solver uses the Gurobi Library \cite{gurobi}.

\subsection{\Tracker{CEM}: Continuous energy minimization}
\label{sec:CEM}
\Tracker{CEM} \cite{Milan:2014:PAMI} formulates the problem in terms of a 
high-dimensional continuous energy. Here, we use the basic approach 
\cite{Andriyenko:2011:CVPR} without explicit occlusion reasoning or 
appearance model. The target state $\mathbf{X}$ is represented by 
continuous $x,y$ coordinates in \emph{all} frames. The energy 
$E(\mathbf{X})$ is made up of several components, including a data term 
to keep the solution close to the observed data (detections), a dynamic 
model to smooth the trajectories, an exclusion term to avoid collisions, 
a persistence term to reduce track fragmentations, and a regularizer.
The resulting energy is highly non-convex and is minimized in an alternating
fashion using conjugate gradient descent and deterministic jump moves.

\subsection{\Tracker{SMOT}: Similar moving objects}
\label{sec:SMOT}
The Similar Multi-Object Tracking (\Tracker{SMOT}) approach 
\cite{Dicle:2013:ICCV} specifically targets situations where target 
appearance is ambiguous and rather concentrates on using the motion as a 
primary cue for data association. Tracklets with similar motion are linked
to longer trajectories using the generalized linear assignment (GLA) 
formulation. The motion similarity and the underlying dynamics of a 
tracklet are modeled as the order of a linear regressor approximating that
tracklet.

\subsection{\Tracker{TBD}: Tracking-by-detection}
\label{sec:TBD}
This two-stage tracking-by-detection (\Tracker{TBD}) approach 
\cite{Geiger:2014:PAMI, Zhang:2013:ICCV} is part of a larger traffic 
scene understanding framework and employs a rather simple data 
association technique. The first stage links overlapping detections
with similar appearance in successive frames into tracklets. The
second stage aims to bridge occlusions of up to 20 frames. Both stages
employ the Hungarian algorithm to optimally solve the matching problem.
Note that we did not re-train this baseline but rather used the original
implementation and parameters provided.

\subsection{\Tracker{SFM}: Social forces for tracking}
\label{sec:SFM}
Most tracking systems work with the assumption that the motion model for each target is independent, but in reality, a pedestrian follows a series of social rules, \ie is subject to social forces according to other moving targets around him/her. These have been defined in what is called the social force model (SFM) \cite{SFM,SFM2} and have recently been applied to multiple people tracking. For the 3D benchmark we include two baselines that include a few hand-designed force terms, such as {\it collision avoidance} or {\it group attraction}. 
The first method (\Tracker{KalmanSFM}) \cite{pellegriniiccv2009} includes those in an online predictive Kalman filter approach while the second (\Tracker{LPSFM}) \cite{Leal-Taixe:2011:ICCVW} includes the social forces in a Linear Programming framework as described in \Sec~\ref{sec:DP_NMS}.
For the 2D benchmark, we include a recent algorithm (\Tracker{MotiCon}) \cite{lealcvpr2014}, which learns an image-based motion context that encodes the pedestrian's reaction to the environment, \ie, other moving objects. The motion context, created from low-level image features, leads to a much richer representation of the physical interactions between targets compared to hand-specified social force models. This allows for a more accurate prediction of the future position of each pedestrian in image space, information that is then included in a Linear Programming framework for multi-target tracking.

\subsection{\Tracker{TC\_ODAL}: Tracklet confidence}
\label{sec:TC_ODAL}

Robust Online Multi-Object Tracking based on Tracklet Confidence and 
Online Discriminative Appearance Learning, or \Tracker{TC\_ODAL} 
\cite{Bae:2014:CVPR}, is the only online method among the baselines. It 
proceeds in two stages. First, close detections are linked to form a set 
of short, reliable tracklets. This so-called local association allows 
one to progressively aggregate confident tracklets. In case of occlusions 
or missed detections, the tracklet confidence value is decreased and a 
global association is employed to bridge longer occlusion gaps. Both 
association techniques are formulated as bipartite matching and tackled 
with the Hungarian algorithm.

Another prominent component of \Tracker{TC\_ODAL} is online appearance learning.
To that end, positive samples are collected from tracklets with high
confidence and incremental linear discriminant analysis (ILDA) is employed
to update the appearance model in an online fashion.

\begin{table*}[tb]
\begin {center}
\setlength{\tabcolsep}{0.15cm}
\begin{tabular}{l| cccccccccccccc}
Method & AvgRank & MOTA & MOTP & FAR & MT(\%) & ML(\%) & FP & FN & IDsw & rel.ID & FM & rel.FM & Hz & Ref.\\
{\bf 2D MOT 2015 } & & & & & & & & & & & & & &\\
\hline
\Tracker{MotiCon} & 9.3 & 23.1 {\tiny $\pm$16.4} & 70.9 & 1.8 & 4.7 & 52.0 & 10404 & 35844 & 1018 & 24.4 & 1061 & 25.5 & 1.4 & \cite{lealcvpr2014} \\
\Tracker{LP2D} & 8.3 & 19.8 {\tiny $\pm$14.2} & 71.2 & 2.0 & 6.7 & 41.2 & 11580 & 36045 & 1649 & 39.9 & 1712 & 41.4 & 112.1 & baseline\\
\Tracker{CEM} & 9.2 & 19.3 {\tiny $\pm$17.5} & 70.7 & 2.5 & 8.5 & 46.5 & 14180 & 34591 & 813 & 18.6 & 1023 & 23.4 & 1.1 & \cite{Milan:2014:PAMI}\\
\Tracker{RMOT} & 10.6 & 18.6 {\tiny $\pm$17.5} & 69.6 & 2.2 & 5.3 & 53.3 & 12473 & 36835 & 684 & 17.1 & 1282 & 32.0 & 7.9 & \cite{yoonwacv2015}\\
\Tracker{SMOT} & 10.7 & 18.2 {\tiny $\pm$10.3} & 71.2 & 1.5 & 2.8 & 54.8 & 8780 & 40310 & 1148 & 33.4 & 2132 & 62.0 & 2.7 &  \cite{Dicle:2013:ICCV}\\
\Tracker{TBD} & 12.3 & 15.9 {\tiny $\pm$17.6} & 70.9 & 2.6 & 6.4 & 47.9 & 14943 & 34777 & 1939 & 44.7 & 1963 & 45.2 & 0.7 & \cite{Geiger:2014:PAMI,Zhang:2013:ICCV}\\
\Tracker{TC\_ODAL} & 12.8 & 15.1 {\tiny $\pm$15.0} & 70.5 & 2.2 & 3.2 & 55.8 & 12970 & 38538 & 637 & 17.1 & 1716 & 46.0 & 1.7 & \cite{Bae:2014:CVPR}\\
\Tracker{DP\_NMS} & 10.7 & 14.5 {\tiny $\pm$13.9} & 70.8 & 2.3 & 6.0 & 40.8 & 13171 & 34814 & 4537 & 104.7 & 3090 & 71.3 & 444.8 & \cite{Pirsiavash:2011:CVPR} \\
 & & & & & & & & & & & & & &\\
{\bf 3D MOT 2015} & & & & & & & & & & & & & &\\
\hline
\Tracker{LPSFM} & 1.7 & 35.9 {\tiny $\pm$06.3} & 54.0 & 2.3 & 13.8 & 21.6 & 2031 & 8206 & 520 & 10.2 & 601 & 11.8 & 8.4 &  \cite{Leal-Taixe:2011:ICCVW}\\
\Tracker{LP3D} & 2.0 & 35.9 {\tiny $\pm$11.1} & 53.3 & 4.0 & 20.9 & 16.4 & 3588 & 6593 & 580 & 9.6 & 659 & 10.9 & 83.5 & baseline\\
\Tracker{KalmanSFM} & 2.3 & 25.0 {\tiny $\pm$08.5} & 53.6 & 3.6 & 6.7 & 14.6 & 3161 & 7599 & 1838 & 33.6 & 1686 & 30.8 & 30.6 & \cite{Pellegrini:2009:ICCV}\\
\end{tabular}
\caption{Quantitative results on all baselines.}
\label{tab:results}
\end{center}
\end{table*}

\section{Evaluation}
\label{sec:evaluation}
Evaluating multiple object tracking to this day remains a surprisingly 
difficult task. Even though many measures have been proposed in the past 
\cite{Smith:2005:CVPRW, Stiefelhagen:2006:CLE, Bernardin:2008:CLE, 
Schuhmacher:2008:ACM, Wu:2006:CVPR, Li:2009:CVPR}, comparing a new 
method against prior art is typically not straightforward. As discussed 
in some of our previous work \cite{Milan:2013:CVPRWS}, the reasons for 
that are diverse ranging from ambiguous ground truth, imprecise metric 
definitions or implementation variations. In this section we will 
describe the entire evaluation procedure of our benchmark in detail.

\captionsetup[subfigure]{labelformat=empty}
\subsection{Annotations}
\label{sec:annotations}
\begin{figure*}
\centering
\subfigure[][ADL-Rundle-1]{
\includegraphics[trim=0cm 0.0cm 0cm 0cm,clip=true,width=0.31\linewidth]{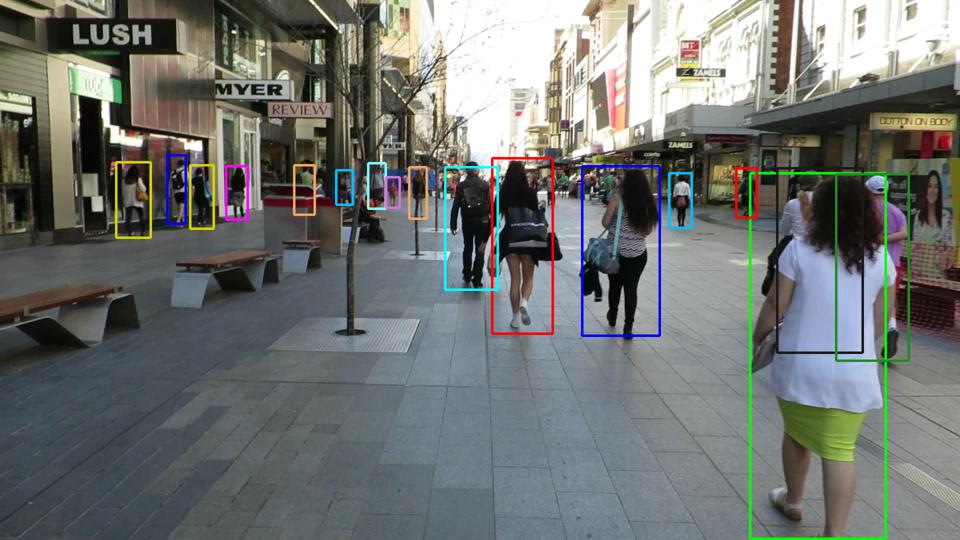} 
\label{fig:adl-rundle-1-gt}}
\subfigure[Venice-2]{
\includegraphics[trim=0cm 0cm 0cm 0cm,clip=true,width=0.31\linewidth]{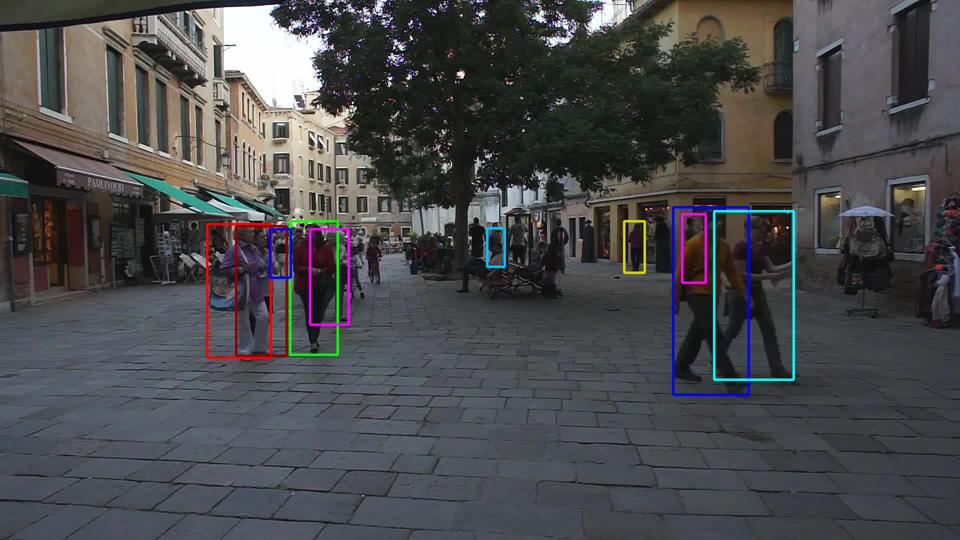} 
\label{fig:venice-2-gt}}
\subfigure[PETS09-S2L2]{
\includegraphics[trim=0cm 3cm 0cm 2cm,clip=true,width=0.31\linewidth]{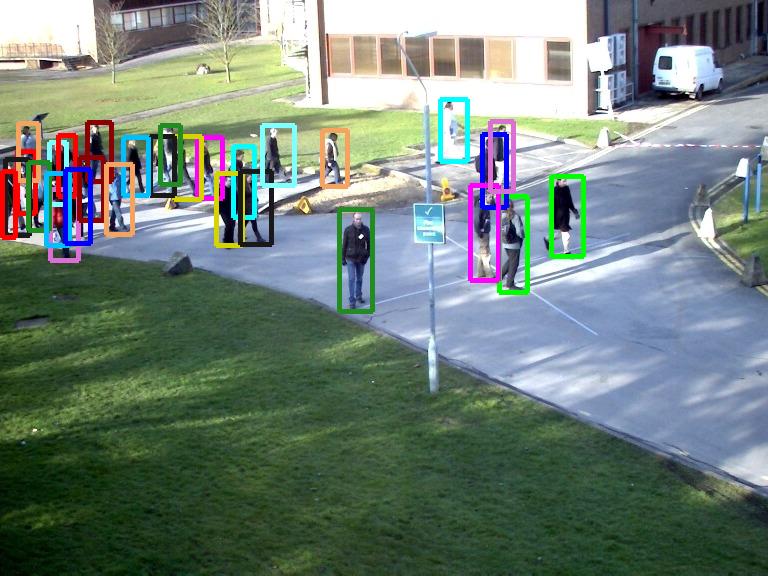} 
\label{fig:pets09-s2l2-gt}}
\caption{Ground truth (manually annotated) bounding boxes on three sequences.}
\label{fig:annotations}
\end{figure*}
As in many other applications, multi-target tracking requires a set of 
labeled (or annotated) videos in order to quantitatively evaluate the 
performance of a particular approach. Unfortunately, human supervision 
is necessary to obtain a reliable set of this, so-called ground truth. 
Depending on factors like object count, image quality, or the 
level-of-detail, annotating video data can be a rather tedious task. 
This is one of the reasons why there exist only relatively few datasets 
with publicly available ground truth.

For the majority of the sequences contained in our benchmark, we employ 
the publicly available ground truth. The 6 new sequences 
(\emph{ADL-Rundle-*} and \emph{Venice-*}) were annotated by us using the 
VATIC annotation tool \cite{Vondrick:2013:IJCV}. We provide the ground 
truth  for the training set, however, to reduce overfitting on unseen 
data, the annotations for the test sequences are withheld.
Annotation samples are illustrated in \Fig~\ref{fig:annotations}.

\subsubsection{Variation in the annotations}

\begin{figure*}[htb]
\centering
\subfigure[][TUD-Campus]{
\includegraphics[height=3.5cm]{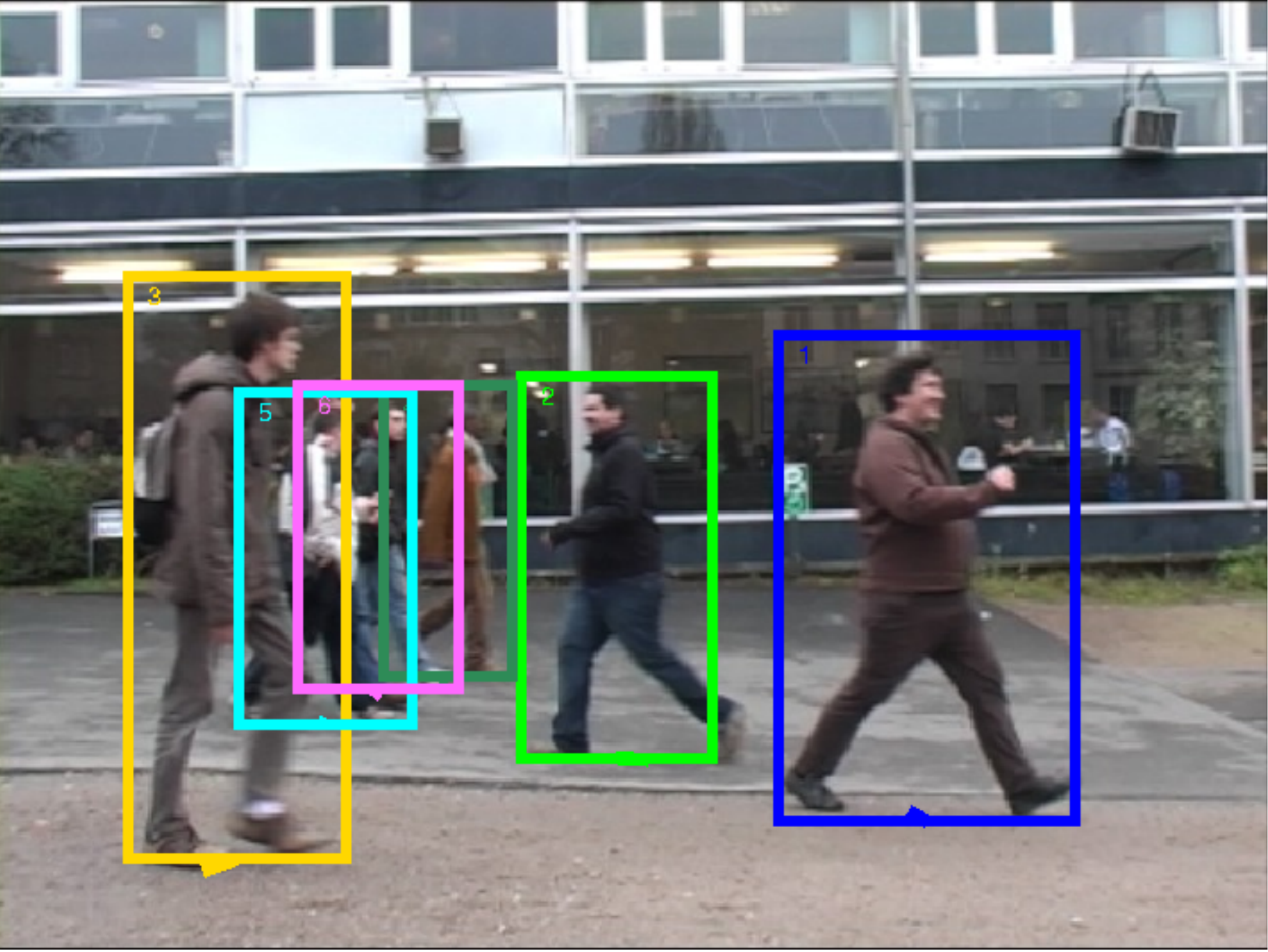} 
\label{fig:tudcampus}}
\subfigure[AVG-TownCentre]{
\includegraphics[height=3.5cm]{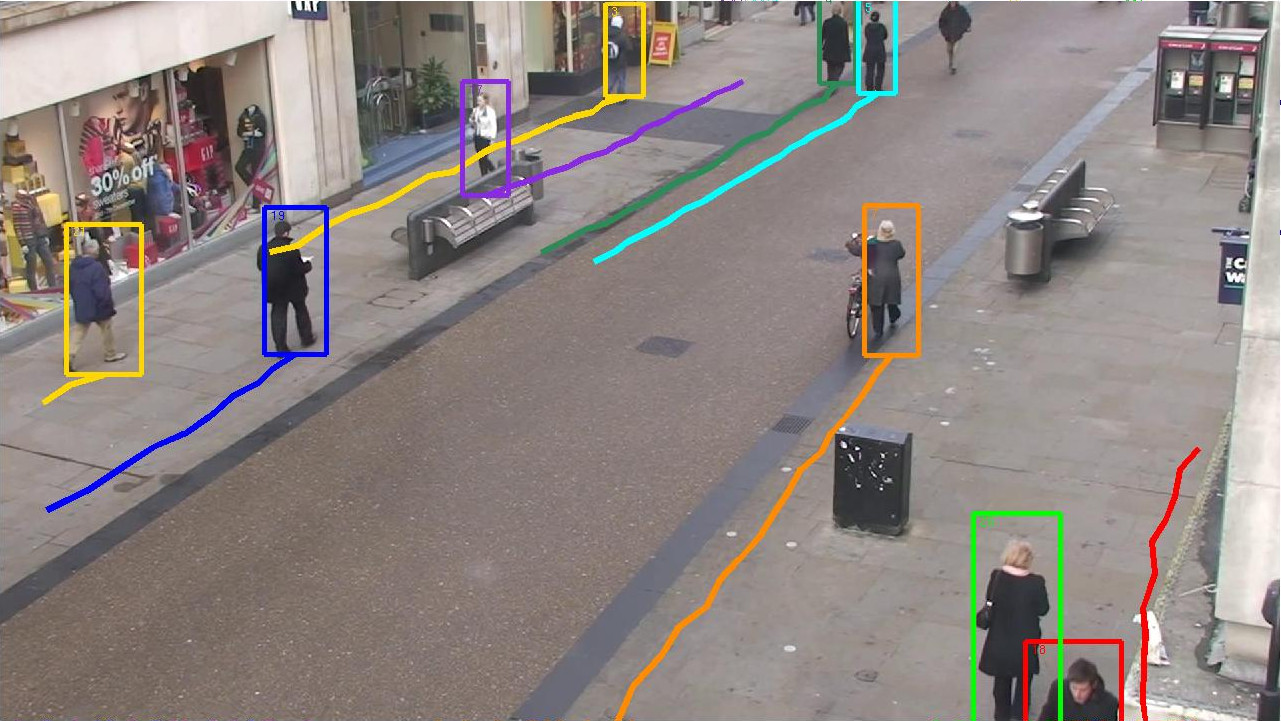} 
\label{fig:avgtowncentre}}
\subfigure[ETH-Crossing]{
\includegraphics[height=3.5cm]{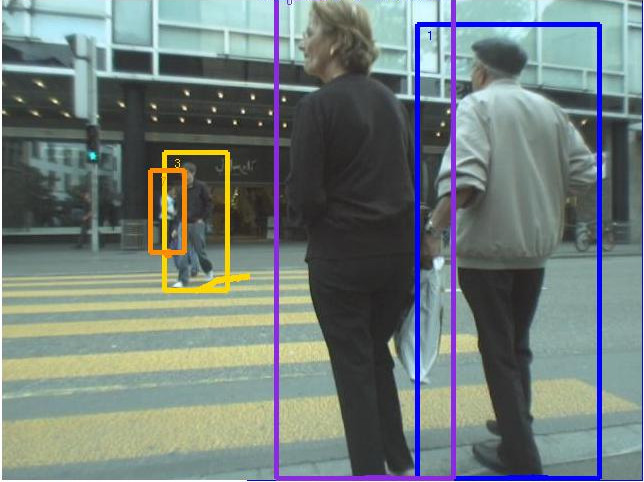} 
\label{fig:ethcrossing}}
\caption{Publicly available ground truth bounding boxes on three sequences.}
\label{fig:annotationrules}
\end{figure*}

Publicly available annotations contain a relatively large amount of
variation.
Since, as of now, we rely on different sources for our annotations, we cannot state that they all follow a set of common rules.  
Some bounding boxes enclose the whole pedestrian, including all the limbs, 
which can lead to bounding boxes that change noticeably in size depending on the pedestrian's pose,
as we can see in \Fig \ref{fig:tudcampus}, blue pedestrian \vs yellow pedestrian.
In \Fig \ref{fig:avgtowncentre}, we see that bounding boxes are not always centered exactly on the pedestrians,
which could cause small shifts in the 3D position estimation. 
Another common problem is that bounding boxes for pedestrians that are close to the camera are usually very tight
around the pedestrian's silhouette compared to pedestrians far away, as we can see in \Fig \ref{fig:ethcrossing}, 
blue bounding box \vs yellow bounding box.
Occlusions are also handled differently among sequences. While some annotations follow pedestrians even under full occlusion \cite{Ferryman:2010:PETS}, 
others create a new trajectory once the pedestrian reappears \cite{Ess:2008:CVPR}.

Recently, a thorough study on face detection benchmarks
\cite{Mathias:2014:ECCV} also showed that annotation policies vary
greatly among sequences and datasets. It also showed that adapting the
evaluation method to be more robust against annotation variation 
plus reannotation of the data with a fixed set of rules changed the
performance of many state-of-the-art methods.

Even though annotations based on a standardized policy are not
available in the benchmark yet, the larger size and stronger variation
in the benchmark already exceed existing benchmarks significantly.
Nonetheless, and following the work in  \cite{Mathias:2014:ECCV}, we commit
to standardizing the set of annotations for all sequences following a common strict set of rules. These annotations will
be published in the second release of the benchmark.

%

\subsection{Evaluation metrics}
\label{sec:evaluation-metrics}
In the past, a large number of metrics for quantitative evaluation of 
multiple target tracking have been proposed \cite{Smith:2005:CVPRW, 
Stiefelhagen:2006:CLE, Bernardin:2008:CLE, Schuhmacher:2008:ACM, 
Wu:2006:CVPR, Li:2009:CVPR}. Choosing ``the right'' one is largely 
application dependent and the quest for a unique, general evaluation metric
is still ongoing. On the one hand, it is desirable to summarize the performance 
into one single number to enable a direct comparison. On the other hand,
one might not want to lose information about the individual errors made
by the algorithms and provide several performance estimates, which 
precludes a clear ranking.

Following the recent trend \cite{Milan:2014:PAMI, Bae:2014:CVPR, 
Wen:2014:CVPR} we employ two sets of measures that have established 
themselves in the literature: The \emph{CLEAR} metrics proposed by 
Stiefelhagen \etal \cite{Stiefelhagen:2006:CLE}, and a set of track 
quality measures introduced by Wu and Nevatia \cite{Wu:2006:CVPR}.
The evaluation scripts used in our benchmark are publicly 
available.\footnote{\url{http://motchallenge.net/devkit}}

\subsubsection{Tracker-to-target assignment}
\label{sec:tracker-assignment}
There are two common prerequisites for quantifying the performance of a 
tracker. One is to determine for each hypothesized output, whether it is a 
true positive (TP) that describes an actual (annotated) target, or 
whether the output is a false alarm (or false positive, FP). This 
decision is typically made by thresholding based on a defined distance 
(or dissimilarity) measure $\dismeas$ (see 
\Sec~\ref{sec:distance-measure}). A target that is missed by any 
hypothesis is a false negative (FN). A good result is expected to have 
as few FPs and FNs as possible. Next to the absolute numbers, we also 
show the false positive ratio measured by the number of false alarms per 
frame (FAF), sometimes also referred to as false positives per image 
(FPPI) in the object detection literature.

\begin{figure*}[t]
\centering
\def\svgwidth{1\linewidth}
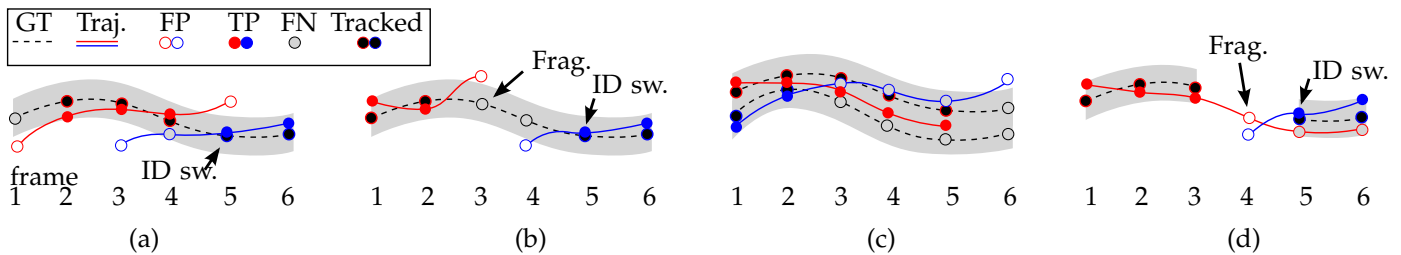
\caption
{
Four cases illustrating tracker-to-target assignments. (a) An ID switch 
occurs when the mapping switches from the previously assigned red track 
to the blue one. (b) A track fragmentation is counted in frame 3 because 
the target is tracked in frames 1-2, then interrupts, and then 
reacquires its `tracked' status at a later point. A new (blue) track hypothesis also 
causes an ID switch at this point. (c) Although the tracking results is 
reasonably good, an optimal single-frame assignment in frame 1 is 
propagated through the sequence, causing 5 missed targets (FN) and 4 
false positives (FP). Note that no fragmentations are counted in frames 
3 and 6 because tracking of those targets is not resumed at a later 
point. (d) A degenerate case illustrating that target re-identification 
is not handled correctly. An interrupted ground truth trajectory will 
typically cause a fragmentation. Also note the less intuitive ID switch, 
which is counted because blue is the closest target in frame 5 that is 
not in conflict with the mapping in frame 4. 
}
\label{fig:mapping}
\end{figure*}

Obviously, it may happen that the same target is covered by multiple 
outputs. The second prerequisite before computing the numbers is then to 
establish the correspondence between all annotated and hypothesized 
objects under the constraint that a true object should be recovered at 
most once, and that one hypothesis cannot account for more than one 
target. 

For the following, we assume that each ground truth trajectory has one 
unique start and one unique end point, \ie that it is not fragmented. 
Note that the current evaluation procedure does not explicitly handle 
target re-identification. In other words, when a target leaves the 
field-of-view and then reappears, it is treated as an unseen target with 
a new ID. As proposed in \cite{Stiefelhagen:2006:CLE}, the optimal 
matching is found using Munkre's (a.k.a.~Hungarian) algorithm. However, 
dealing with video data, this matching is not performed independently 
for each frame, but rather considering a temporal correspondence.
More precisely, if a ground truth object $i$ is matched to hypothesis 
$j$ at time $t-1$ \emph{and} the distance (or dissimilarity) between $i$ 
and $j$ in frame $t$ is below $\simthresh$, then the correspondence 
between $i$ and $j$ is carried over to frame $t$ even if there exists another
hypothesis that is closer to the actual target. A mismatch error (or 
equivalently an identity switch, IDSW) is counted if a ground truth 
target $i$ is matched to track $j$ and the last known assignment was $k 
\ne j$. Note that this definition of ID switches is more similar to 
\cite{Li:2009:CVPR} and stricter than the original one 
\cite{Stiefelhagen:2006:CLE}. Also note that, while it is certainly 
desirable to  keep the number of ID switches low, their absolute number 
alone is not always expressive to assess the overall performance, but 
should rather be considered in relation to the number of recovered 
target. The intuition is that a method that finds twice as many 
trajectories will almost certainly produce more identity switches. For 
that reason, we also state the relative number of ID switches, which is 
computed as IDSW / Recall.

These relationships are illustrated in \Fig~\ref{fig:mapping}. For 
simplicity, we plot ground truth trajectories with dashed curves, and 
the tracker output with solid ones, where the color represents a unique 
target ID. The grey areas indicate the matching threshold (see next 
section). Each true target that has been successfully recovered in one 
particular frame is represented with a filled black dot with a stroke 
color corresponding to its matched hypothesis. False positives and false 
negatives are plotted as empty circles. See figure caption for more 
details.

After determining true matches and establishing the correspondences it
is now possible to compute the metrics. We do so by concatenating all
test sequences and evaluating on the entire benchmark. This is in
general more meaningful instead of averaging per-sequences figures due to
the large variation in the number of targets.

\subsubsection{Distance measure}
\label{sec:distance-measure}
\begin{figure}[t]
\centering
\includegraphics[width=.8\linewidth]{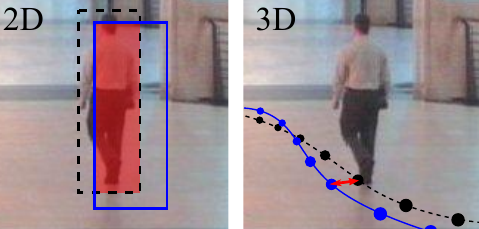}
\caption
{
The closeness between the tracker output (blue) and the true location 
of a target (black dashed) can be computed as a bounding box overlap or 
as Euclidean distance in world coordinates.
}
\label{fig:match-distance}
\end{figure}
To determine how close a tracker hypothesis is to the actual target, we 
will distinguish two cases as described below (see also 
\Fig~\ref{fig:match-distance}).

{\bf{2D.}}
In the most general case, the relationship between ground truth objects 
and a tracker output is established using bounding boxes on the image 
plane. Similar to object detection \cite{Everingham:2012:VOC}, the 
intersection over union (a.k.a. the Jaccard index) is usually employed 
as the similarity criterion, while the threshold $\simthresh$ is set to 
$0.5$ or $50\%$.

{\bf{3D.}}
When both locations, that of the tracker and that of the ground truth, 
are available as points in world coordinates, it is more sensible to 
directly compute the performance in 3D. To that end, $\dismeas$ simply 
corresponds to the Euclidean distance and $\simthresh$ is set to $1$ 
meter for pedestrian tracking.

\subsubsection{Multiple Object Tracking Accuracy}
\label{sec:mota}
The MOTA \cite{Stiefelhagen:2006:CLE} is perhaps the most widely used 
figure to evaluate a tracker's performance. The main reason for this is 
its expressiveness as it combines three sources of errors defined above:
\begin{equation}
\text{MOTA} = 
1 - \frac
{\sum_t{(\text{FN}_t + \text{FP}_t + \text{IDSW}_t})}
{\sum_t{\text{GT}_t}},
\label{eq:mota}
\end{equation}
where $t$ is the frame index and GT is the number of ground truth 
objects. We report the percentage MOTA $(-\infty, 100]$ in our 
benchmark. Note that MOTA can also be negative in cases where the number 
of errors made by the tracker exceeds the number of all objects in the 
scene.

Even though the MOTA score gives a good indication of the overall 
performance, it is highly debatable whether this number alone can serve 
as a single performance measure. 

{\bf{Robustness.}}
One incentive behind compiling this benchmark was to reduce dataset bias
by keeping the data as diverse as possible. The main motivation is to
challenge state-of-the-art approaches and analyze their performance in
unconstrained environments and on unseen data. Our experience shows that
most methods can be heavily overfitted on one particular dataset, but are not general enough to handle an entirely
different setting without a major change in parameters or even in the
model.

To indicate the robustness of each tracker over \emph{all} benchmark
sequences, we show the standard deviation on their MOTA score.

\subsubsection{Multiple Object Tracking Precision}
\label{sec:motp}

The Multiple Object Tracking Precision is the average dissimilarity 
between all true positives and their corresponding ground truth targets. 
For bounding box overlap, this is computed as 
\begin{equation}
\text{MOTP} = 
\frac
{\sum_{t,i}{d_{t,i}}}
{\sum_t{c_t}},
\label{eq:motp}
\end{equation}
where $c_t$ denotes the number of matches in frame $t$ and $d_{t,i}$ is 
the bounding box overlap of target $i$ with its assigned ground truth 
object. MOTP thereby gives the average overlap between all correctly 
matched hypotheses and their respective objects and ranges between 
$\simthresh := 50\%$ and $100\%$.

It is important to point out that MOTP is a 
measure of localization precision, \emph{not} to be confused with the 
\emph{positive predictive value} or \emph{relevance} in the context of 
precision / recall curves used, \eg, in object detection.

As we can see in \Tab~\ref{tab:results}, MOTP shows a remarkably low 
variation across different methods ranging between $69.6\%$ and $71.6\%$. 
The main reason for this is that this localization measure is primarily 
dominated by the detections and the annotations and is less influenced 
by the actual tracker output.

If computed in 3D, the definition changes slightly to 
\begin{equation}
\text{MOTP}_{\text{3D}} = 
1-\frac
{\sum_{t,i}{d_{t,i}}}
{\simthresh \cdot \sum_t{c_t}}.
\label{eq:motp3d}
\end{equation}
Note that here it is normalized to be between 0 and 100\%.

\subsubsection{Track quality measures}
\label{sec:track-measures}


Each ground truth trajectory can be classified as mostly tracked (MT), 
partially tracked (PT), and mostly lost (ML). This is done based on how 
much of the trajectory is recovered by the tracking algorithm. A target 
is mostly tracked if it is successfully tracked for at least $80\%$ of 
its life span. Note that it is irrelevant for this measure whether the 
ID remains the same throughout the track. If a track is only recovered 
for less than $20\%$ of its total length, it is said to be mostly lost 
(ML). All other tracks are partially tracked. A higher number of MT and 
few ML is desirable. We report MT and ML as a ratio of mostly tracked 
and mostly lost targets to the total number of ground truth 
trajectories.

In certain situations one might be interested in obtaining long, 
persistent tracks without gaps of untracked periods. To that end, the 
number of track fragmentations (FM) counts how many times a ground truth 
trajectory is interrupted (untracked). In other words, a fragmentation 
is counted each time a trajectory changes its status from tracked to 
untracked and tracking of that same trajectory is resumed at a later 
point. Similarly to the ID switch ratio 
(\cf~\Sec~\ref{sec:tracker-assignment}), we also provide the relative 
number of fragmentations as FM / Recall.

\subsubsection{Runtime}
\label{sec:runtime}
\begin{figure}[t]
\centering
\includegraphics[width=1\linewidth]{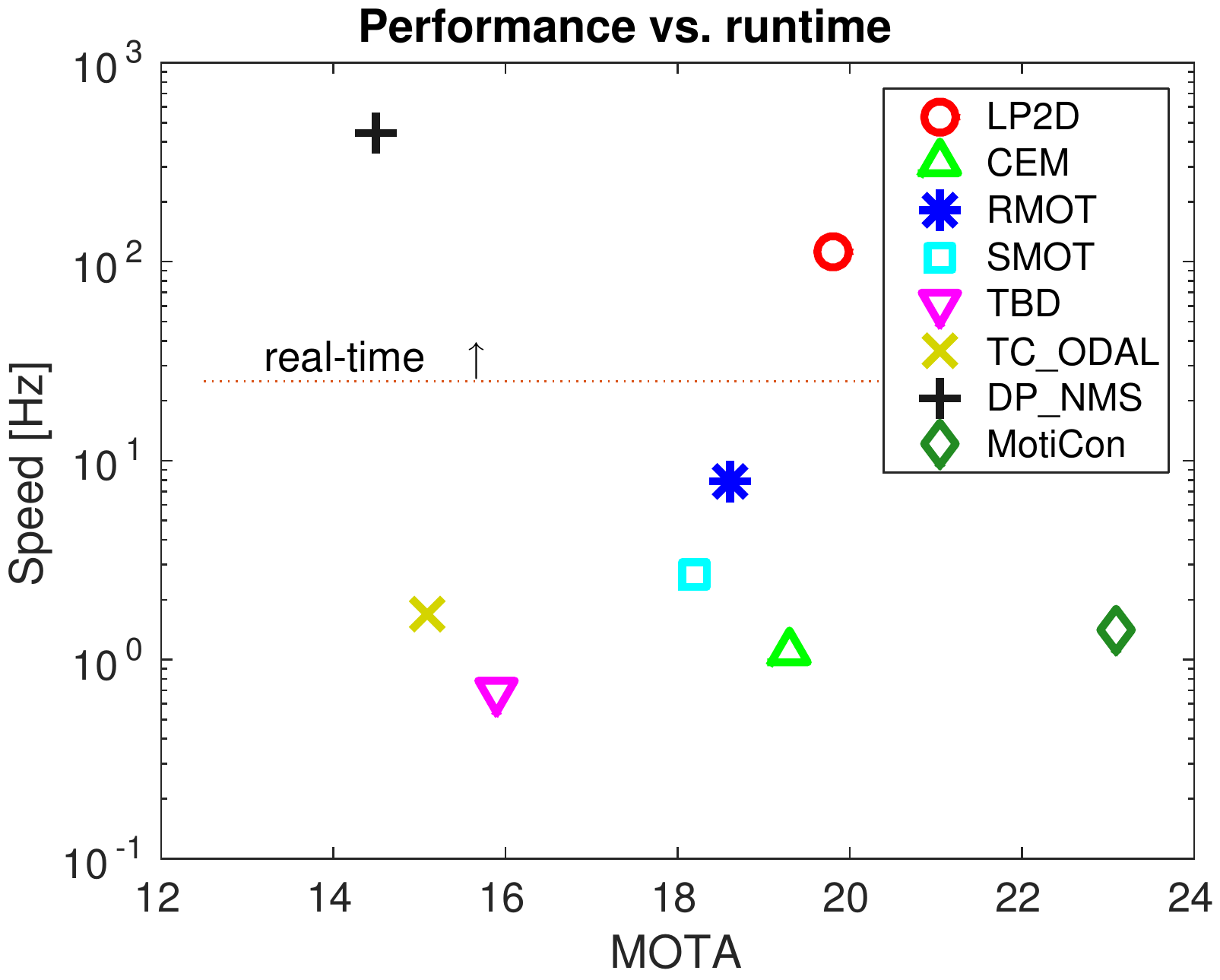}
\caption
{
Each marker represents a tracker's performance measured by MOTA 
($x$-axis) and its speed measured in frames per second (FPS) [Hz],
\ie higher and more right is better. Real-time ability is assumed at 
25 FPS.
}
\label{fig:MOTAvsSpeed}
\end{figure}
Most research in multi-target tracking focuses on pushing the 
performance towards more accurate results with fewer errors. However, 
from the practical point of view, a method should be able to compute the 
results in a reasonable time frame. Of course, `reasonable' varies  
depending on the application. Autonomous vehicles or tasks in 
robotics would require real-time functionality; surveillance assistance 
may tolerate a certain delay, while long-term video analysis may allow 
even much longer processing times. We demonstrate the relationship 
between tracker accuracy and speed in \Fig~\ref{fig:MOTAvsSpeed}. 
As we can see, the fastest approach, \Tracker{DP\_NMS},  performs worse on average while the baseline
\Tracker{LP2D} provides a good balance between speed and performance.

Note that accurately measuring the efficiency of each method is not 
straightforward. All baselines from \Sec~\ref{sec:baselines} were 
executed on the same hardware ($2.6$ GHz $\times 16$ cores CPU, $32$ GB 
RAM) and the reported numbers do not include the detector's time. For 
all submitted results we cannot verify the efficiency ourselves and 
therefore report the runtime as specified by the respective user.

\subsubsection{Tracker ranking}
\label{sec:ranking}
As we have seen in this section, there are a number of reasonable 
performance measures to assess the quality of a tracking system, which 
makes it rather difficult to reduce the evaluation to one single number. 
To nevertheless give an intuition on how each tracker performs compared 
to its competitors, we compute and show the average rank for each one by 
ranking all trackers according to each metric and then averaging across 
all ten performance measures.
Interestingly, the average rank roughly corresponds to the MOTA ordering,
which indicates that the tracking accuracy is a good approximation of the
overall tracker performance.


  \section{Conclusion and Future Work}
 \label{sec:conclusion}
 
We have presented a novel platform for evaluating multi-target 
tracking approaches. Our centralized benchmark consists of both
existing public videos as well as new challenging sequences and
is open for new submissions. We believe that this will enable
a fairer comparison and guide research towards developing
more generic methods that perform well in unconstrained
environments and on unseen data.

In future, we will work on the standardization of the annotations for all sequences, 
continue our workshops and challenges series, 
and also introduce various other (sub-)benchmarks to welcome researchers
and practitioners from other disciplines.

\ifCLASSOPTIONcaptionsoff
  \newpage
\fi

\bibliographystyle{ieee}
\bibliography{refs-lau,refs-short,refs-anton,refs-new}


\end{document}